%% file: main.tex
\pdfoutput=1

\documentclass[11pt]{article}

\usepackage[final]{acl}

\usepackage{times}
\usepackage{latexsym}

\usepackage[T1]{fontenc}

\usepackage[utf8]{inputenc}

\usepackage{microtype}

\usepackage{inconsolata}

\usepackage{graphicx}

\usepackage[english]{babel}

\usepackage{multirow}
\usepackage{multicol}
\usepackage{amsmath}
\usepackage{booktabs}
\usepackage{amsthm}
\usepackage{amssymb}
\usepackage{makecell}
\usepackage{subcaption}
\usepackage{arydshln}
\usepackage{hyperref}
\usepackage{pifont}
\usepackage{tcolorbox}
\usepackage{bm}
\usepackage{mathtools}
\usepackage{xr}
\usepackage{bbding}
\usepackage{ulem}
\usepackage{tabularx,ragged2e}

\newcommand\mycdots{\hbox to 0.75em{$\cdot$\hss$\cdot$\hss$\cdot$}}

\newcommand{\NA}{\rule[0.5ex]{0.7em}{0.5pt}}

\title{Astra: Efficient Transformer Architecture and Contrastive Dynamics Learning for Embodied Instruction Following}

\author{
Yueen Ma\textsuperscript{\rm 1}, 
Dafeng Chi\textsuperscript{\rm 2}, 
Shiguang Wu\textsuperscript{\rm 2}, 
Yuecheng Liu\textsuperscript{\rm 2}, 
Yuzheng Zhuang\textsuperscript{\rm 2}, 
Irwin King\textsuperscript{\rm 1}\\
Department of Computer Science and Engineering, The Chinese University of Hong Kong\textsuperscript{\rm 1} \\
Huawei Noah's Ark Lab\textsuperscript{\rm 2}\\
\texttt{\{yema21, king\}@cse.cuhk.edu.hk}\\
\texttt{\{chidafeng1, wushiguang, liuyuecheng1, zhuangyuzheng\}@huawei.com}\\
}

\begin{document}
\maketitle
\begin{abstract}
Vision-language-action models have gained significant attention for their ability to model multimodal sequences in embodied instruction following tasks. However, most existing models rely on causal attention, which we find suboptimal for processing sequences composed of interleaved segments from different modalities. In this paper, we introduce \mbox{Astra}\footnote{\url{https://github.com/yueen-ma/Astra}}, a novel Transformer architecture featuring trajectory attention and learnable action queries, designed to efficiently process segmented multimodal trajectories and predict actions for imitation learning. Furthermore, we propose a contrastive dynamics learning objective to enhance the model's understanding of environment dynamics and multimodal alignment, complementing the primary behavior cloning objective. Through extensive experiments on three large-scale robot manipulation benchmarks, Astra demonstrates substantial performance improvements over previous models.
\end{abstract}

\input{sections/1_intro}
\input{sections/2_related}
\input{sections/3_method}

\input{sections/4_experiments}

\section{Conclusion}

This paper introduces Astra, an efficient Transformer architecture designed for multimodal trajectories in embodied instruction following tasks. Astra distinguishes itself from standard Transformer decoders through two novel components: trajectory attention and action queries. Trajectory attention harnesses the unique characteristics of multimodal EIF trajectories, facilitating enhanced information flow among tokens within each segment. Combined with our action queries, which enable parallel information extraction for individual dimensions, Astra achieves segment-level decoding. Furthermore, we incorporate a contrastive dynamics learning objective to explicitly train the model to learn environment dynamics, which also improves multimodal alignment. This further elevates Astra's performance in imitation learning. Comprehensive experiments across three large-scale benchmarks demonstrate substantial performance gains by Astra. Detailed ablation studies and qualitative analyses further validate its effectiveness.

\section*{Limitations}
Although real-world robot evaluation is a common practice, there is no widely accepted real-world benchmark for embodied instruction following tasks due to the difficulty of precisely replicating environmental setups across different institutions. Benchmarks based on simulated environments offer the advantage of highly controllable settings, eliminating variable factors that can lead to imprecise measurements. Therefore, we focus on well-established simulated benchmarks that evaluate various aspects of model performance and generalization. Because 3D information can be more informative than 2D image inputs for EIF tasks, Astra can also be extended to integrate 3D vision modules to further improve performance, as the Astra backbone is compatible with various vision, language, or action modules.

\section*{Acknowledgements}
The work described in this paper was partially supported by the Research Grants Council of the Hong Kong Special Administrative Region, China (CUHK 2410072, RGC R1015-23) and Huawei Technology Investment, Limited (CUHK 6906061).


\input{main.bbl}
\clearpage

\appendix



\input{sections/appendix}

\end{document}

%% file: sections/1_intro.tex
\section{Introduction}

Vision-language-action models (VLAs) \cite{DBLP:journals/corr/abs-2307-15818} have recently emerged to address embodied instruction following tasks (EIF) \cite{DBLP:conf/iclr/LuWLLT25}. Previous multimodal models, such as vision-language models (VLMs), have demonstrated proficiency in handling both visual and textual inputs, successfully tackling a variety of tasks, such as visual question answering and image captioning \cite{DBLP:journals/pami/ZhangHJL24}. In contrast, VLAs differ from VLMs in that they can interpret language instructions, visually perceive their environment, and execute actions to fulfill specified embodied tasks. As a result, VLAs can empower embodied agents to interact with the physical world.

\begin{figure}[t]
    \centering
    \begin{subfigure}[b]{0.49\columnwidth}
        \centering
        \includegraphics[width=0.7\textwidth, trim={0.625in 9.38in 6.5in 0.375in},clip]{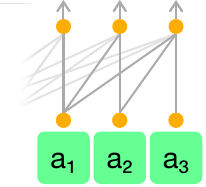}
        \caption{Causal attention.}
    \end{subfigure}
    \hfill
    \begin{subfigure}[b]{0.49\columnwidth}
        \centering
        \includegraphics[width=0.7\textwidth, trim={0.625in 7.63in 6.5in 2.125in},clip]{figures/intro.pdf}
        \caption{Trajectory attention.}
    \end{subfigure}
    \caption{Comparison of information flow in an action segment. Squares represent tokens, while orange dots represent their embeddings. Three action tokens comprise an action ``segment''. The lines illustrate information flow from input embeddings (bottom) to output embeddings (top) through a Transformer self-attention layer. In trajectory attention, tokens attend not only to preceding tokens, as in causal attention, but also to subsequent tokens within the same segment, as indicated by the green lines.}
    \label{fig:intro}
\end{figure}

\begin{figure*}
 \centering
    \includegraphics[width=1.0\textwidth, trim={0in 7.59in 0in 0.4in},clip]{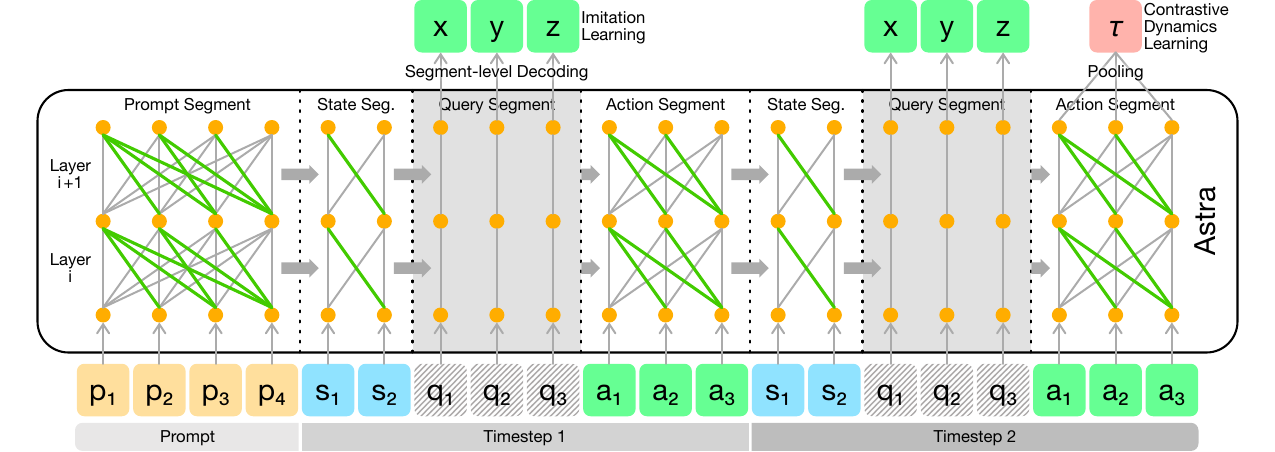}
    \caption{The architecture of Astra. A trajectory $\tau$ comprises a prompt segment $p_{1:4}$, state segments $s_{1:2,t}$, action segments $a_{1:3,t}$. Learnable action queries $q_{1:3,t}$ are inserted after state segments to extract information for action generation. Vertical dashed lines separate these segments. Token embeddings (orange dots) can attend to embeddings in all previous segments (thick horizontal arrows) and to all embeddings within the same segment (gray and green lines). Notably, action queries are hidden from other tokens and can only read from preceding tokens. To facilitate contrastive dynamics learning, Astra can also encode the entire trajectory by pooling the embeddings of the last segment (red box).
    }
    \label{fig:overview}
\end{figure*}

To accommodate multimodal inputs, previous Transformer-based VLMs \cite{DBLP:journals/corr/abs-2404-07214} explored designing special types of self-attention to better suit the unique properties of different modalities. For example, in the task of image captioning, causal attention is not ideal for encoding images because there is no inherent causal relationship among image patches \cite{DBLP:conf/icml/0008LSH23}. Consequently, these VLMs allow bidirectional attention for image tokens while maintaining causal attention for text tokens.

We have similarly observed that multimodal sequences in EIF tasks, which are often referred to as trajectories, exhibit unique properties that can be more effectively captured by a novel type of self-attention, named trajectory attention, as illustrated in Figure~\ref{fig:intro}~\&~\ref{fig:matrix_comparison}. Specifically, each language prompt, state, or action consists of multiple tokens, which we collectively refer to as a ``segment.'' For instance, embodied agents often utilize multiple camera angles, resulting in a state that comprises a segment of tokens, with each token corresponding to a different camera. These state tokens lack causal relationships within the same segment, as they are conditionally independent. The same holds for action tokens that correspond to action dimensions. Therefore, causal attention hinders full information flow within a segment because tokens are restricted from attending to subsequent tokens.

To overcome this limitation, we design trajectory attention with two key characteristics: inter-segment attention is causal, and intra-segment attention is bidirectional. Since a VLA model only needs to encode the language prompt and follow the corresponding instruction, we also apply bidirectional attention to the prompt segment. Consequently, our model processes EIF trajectories at the segment level. Accordingly, we also devise a segment-level decoding scheme that generates a segment as a whole. Drawing inspiration from DETR's object query \cite{DBLP:conf/eccv/CarionMSUKZ20, DBLP:journals/mta/ChenLZT24} for object detection, we employ one learnable action query for each action dimension. Each action query extracts the most relevant information for its corresponding action dimension from preceding tokens and generates the optimal action independently of other action queries. By combining trajectory attention and action queries, we introduce an efficient Transformer architecture for EIF trajectories, which we name the \underline{\textbf{A}}ction-predicting \underline{\textbf{S}}egment-level \underline{\textbf{Tra}}nsformer, or \textbf{Astra} for short. Figure~\ref{fig:overview} provides an overview of Astra.

The Astra architecture also possesses the capability to encode the entire sequence, which opens the possibility for contrastive dynamics learning (CDL), as shown in Figure~\ref{fig:overview}. Numerous prior approaches have explored incorporating dynamics learning to bolster the main imitation learning task \cite{DBLP:journals/corr/abs-2402-02385}, but these efforts typically rely on decoding tasks: forward dynamics methods aim to predict the next state, while inverse dynamics methods attempt to reconstruct the action between two consecutive states. Such approaches often add considerable model complexity, such as requiring a video generator.

Our CDL objective instead leverages the encoding capabilities of Astra. As illustrated in Figure~\ref{fig:cdl}, we create positive samples using a novel action perturbation technique to augment action segments. Negative samples are constructed by mismatching segments with those from other trajectories, thereby violating the environment dynamics. By distinguishing positive samples from negative ones, it learns the correct dynamics, which in turn enhances performance on downstream EIF tasks. Due to the encoding nature of CDL, its implementation simply requires a classification head consisting of a pooling layer followed by a linear layer---a significantly lighter overhead compared to previous decoding-based dynamics learning methods. From another perspective, CDL also serves as a representation learning approach for multimodal alignment \cite{DBLP:journals/ijon/XiaoLWZQJHC25}, as it requires effectively encoding the multimodal trajectories.

The main contributions of this paper are:
\begin{itemize}
    \item We introduce Astra, an efficient Transformer architecture featuring trajectory attention and action queries, designed to efficiently process multimodal trajectories on the segment level;
    \item We propose a contrastive dynamics learning objective that enhances Astra's understanding of environment dynamics and its multimodal encoding capabilities, thereby complementing imitation learning;
    \item Extensive experiments across three large-scale robot manipulation benchmarks demonstrate that Astra significantly outperforms state-of-the-art methods, showcasing the effectiveness of our approach.
\end{itemize}

%% file: sections/2_related.tex
\section{Related Work}


\paragraph{Vision-Language-Action Models.} VLAs \cite{DBLP:journals/corr/abs-2405-14093} are a new class of multimodal models designed to generate actions based on specified language prompts and perceived environments, first proposed by RT-2 \cite{DBLP:journals/corr/abs-2307-15818}. These models adapt pretrained large VLMs to predict actions for EIF tasks and are often referred to as ``large VLAs.'' Representative works include RT-2 \cite{DBLP:conf/icra/ONeillRMGPLPGMJ24}, OpenVLA \cite{DBLP:journals/corr/abs-2406-09246}, and $\pi_0$ \cite{DBLP:journals/corr/abs-2410-24164}. 

Another line of work does not utilize pretrained VLMs and instead builds VLAs from scratch, which are termed ``generalized VLAs.'' These models predominantly draw upon the pioneering foundations laid by DT \cite{DBLP:conf/nips/ChenLRLGLASM21} and TT \cite{DBLP:conf/nips/JannerLL21}, where reinforcement learning is framed as sequence modeling problems. Gato \cite{DBLP:journals/tmlr/ReedZPCNBGSKSEBREHCHVBF22} explored the use of a single Transformer model \cite{DBLP:conf/nips/VaswaniSPUJGKP17} for tasks spanning various domains. RT-1 \cite{DBLP:conf/rss/BrohanBCCDFGHHH23} was the first robotics Transformer. VIMA \cite{DBLP:journals/corr/abs-2210-03094} studied multimodal prompts. Astra can also be categorized as a generalized VLA. However, distinct from these prior VLA models, which rely on causal or cross attention mechanisms, we propose a more efficient Transformer architecture for multimodal EIF tasks.

\paragraph{Multimodal Transformers \& Learnable Queries.} Several VLMs \cite{DBLP:journals/corr/abs-2404-07214}, such as UniLM \cite{DBLP:conf/nips/00040WWLWGZH19}, M6 \cite{DBLP:journals/corr/abs-2103-00823}, and PaliGemma \cite{DBLP:journals/corr/abs-2407-07726} have endeavored to optimize Transformer's self-attention for vision-language inputs by proposing various attention types, such as block attention. To the best of our knowledge, our architecture is the first VLA designed to accommodate multimodal EIF trajectories with a unique self-attention mechanism. 

First introduced in DETR \cite{DBLP:journals/mta/ChenLZT24, DBLP:conf/eccv/CarionMSUKZ20}, learnable object queries have shown promising results in extracting information for object detection. BLIP-2 \cite{DBLP:conf/icml/0008LSH23} used a similar strategy to extract visual embeddings for vision-language tasks. In our approach, we employ learnable action queries at the action-dimension level to extract information most relevant to individual action dimensions.

\paragraph{Dynamics Learning \& Multimodal Contrastive Learning.} Many recent dynamics learning approaches \cite{DBLP:conf/icml/LiGJ0HSSGW24, DBLP:conf/iclr/SunMMBHK23, DBLP:conf/nips/Liu0GA22} can be classified into two categories: forward dynamics learning and inverse dynamics learning. Most of these methods rely on extra generative modules, such as video generators \cite{DBLP:journals/corr/abs-2410-06158, DBLP:conf/nips/DuY0DN0SA23}. Our CDL leverages contrastive learning and involves only an encoding process using a lightweight linear head.

A series of VLMs have demonstrated the significance of contrastive learning in enhancing multimodal interaction \cite{DBLP:journals/pami/ZhangHJL24}. However, contrastive learning methods for EIF trajectories, such as R3M \cite{DBLP:conf/corl/NairRKF022} and VIP \cite{DBLP:conf/iclr/MaSJBK023}, primarily focus on improving visual representations. In contrast, our CDL task compels the model to align all three modalities, thereby enabling more effective encoding of EIF trajectories.

%% file: sections/3_method.tex
\section{Method}

\begin{figure*}
    \centering
    \centering
    \begin{subfigure}[b]{0.49\textwidth}
        \centering
        \includegraphics[width=0.8\textwidth, trim={0.5in 6.25in 3.75in 0.5in},clip]{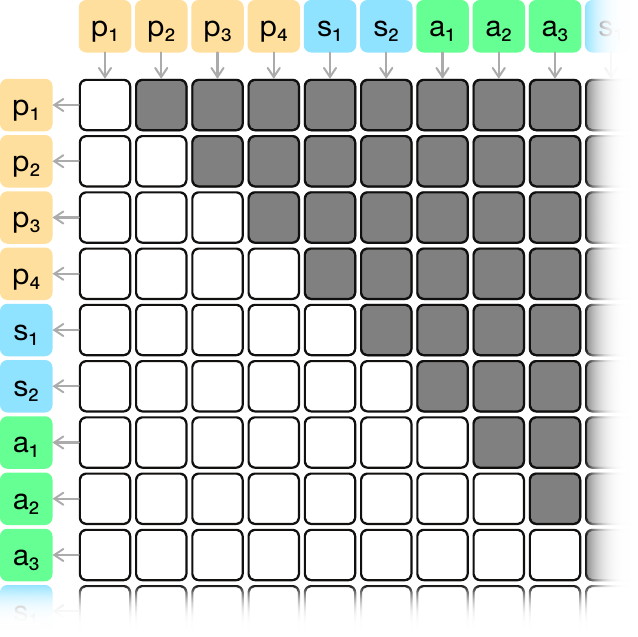}
        \caption{Causal attention.}
    \end{subfigure}
    \hfill
    \begin{subfigure}[b]{0.49\textwidth}
        \centering
        \includegraphics[width=0.8\textwidth, trim={0.5in 6.25in 3.75in 0.5in},clip]{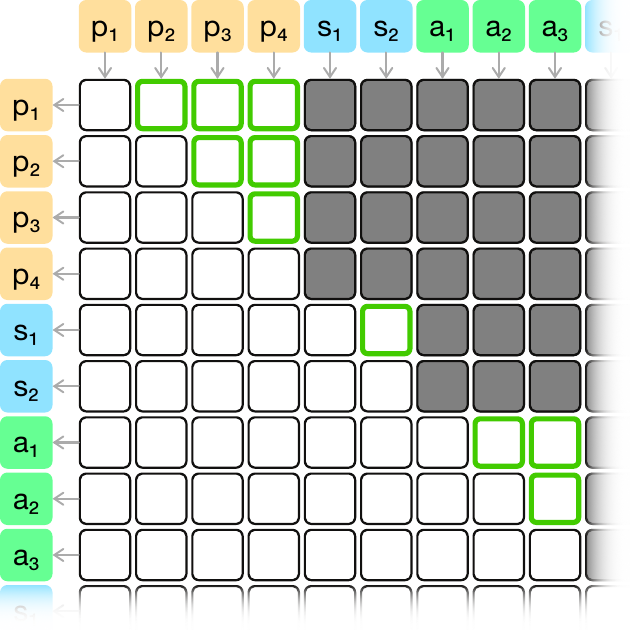}
        \caption{Trajectory attention.}
    \end{subfigure}
    \caption{Attention matrices of causal and trajectory attention. The direction of attention is from the top (input) to the left (output). Dark cells represent attention masks. Green-bordered cells highlight additional information flow enabled by trajectory attention, corresponding to the green lines in Figure~\ref{fig:intro}.}
    \label{fig:matrix_comparison}
\end{figure*}

\subsection{Preliminaries}
Multimodal sequences in embodied instruction following tasks are often referred to as trajectories \cite{DBLP:conf/iclr/LuWLLT25}. These trajectories consist of a language instruction ($p$), states ($s$), and actions ($a$). An trajectory is denoted as $\tau = (p, s_{t=1}, a_{t=1}, \dots, s_{t=T}, a_{t=T})$, where $t$ represents the timestep. Each element in the trajectory---$p$, $s_{t}$, or $a_{t}$---comprises a segment of tokens. For instance, a state $s_t$ is a segment $s_{1:M, t} = (s_{1, t}, s_{2, t}, \dots, s_{M, t})$, where each element is a token representing an image from a particular camera angle. Action tokens in $a_t$ represent either $SE(2)$ actions or 6D pose actions. Tokens in $p$ are standard language tokens. Therefore, a trajectory at the token level can be written as $\tau = (p_{1:L}, s_{1:M, t=1}, a_{1:N, t=1}, \dots, s_{1:M, t=T}, a_{1:N, t=T})$. $L$, $M$, and $N$ are the length of their corresponding segments. The goal is to obtain a policy that can generate an optimal action based on the past trajectory, expressed as $\pi(a_t | p, s_{\leq t}, a_{< t})$.

\subsection{Architecture of Astra}

Astra consists of two main novel components: trajectory attention and action queries, as illustrated in Figure~\ref{fig:overview}. Most VLMs \cite{DBLP:journals/corr/abs-2502-13923} utilize Transformer decoders as the backbone for natural language generation (NLG). To ensure that future tokens are not visible, these Transformer decoders typically use causal attention. Previous VLAs \cite{DBLP:conf/icra/ONeillRMGPLPGMJ24, DBLP:conf/rss/BrohanBCCDFGHHH23} have also adopted this approach for action generation. Although causal attention is well-suited for NLG, where language tokens are generated autoregressively, it is not the optimal attention mechanism for modeling multimodal trajectories in EIF tasks.

\begin{figure*}
    \centering
     \begin{subfigure}[b]{0.22\textwidth}
        \centering
            \includegraphics[width=1.0\textwidth, trim={0.5in 9.625in 6.875in 0.5in},clip]{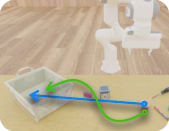}
        \caption{Action perturbation.}
        \label{fig:action_perturb}
     \end{subfigure}
    \hfill 
    {\color{gray} \vrule width 1.0pt height 0.135\textheight} 
    \hfill 
     \begin{subfigure}[b]{0.66\textwidth}
        \centering
            \includegraphics[width=1.0\textwidth, trim={1in 9.075in 2.25in 0.55in},clip]{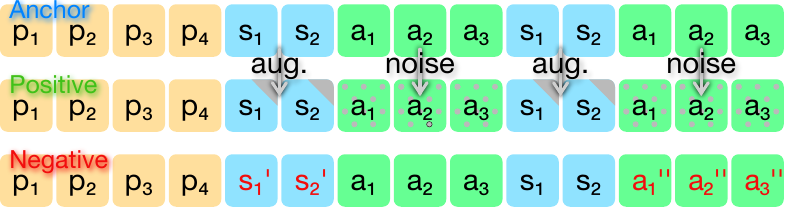}
        \caption{Construction of positive and negative samples.}
        \label{fig:pos_neg}
     \end{subfigure}
    \caption{Contrastive dynamics learning. (a) In the anchor trajectory (blue arrow), the object on the right is picked up and placed into the bin on the left. A slightly deviated trajectory (green arrow) can still reach the desired destination, enabling action perturbation to be used in constructing positive samples. (b) Given the anchor, we construct a positive sample by applying image augmentation (aug.) and the proposed action perturbation. Negative samples are created by mismatching states and actions from other trajectories.
    }
    \label{fig:cdl}
\end{figure*}

\paragraph{Trajectory attention.} Images of the state $s_t$ from multiple cameras arrive simultaneously, lacking causality among themselves. They are determined solely by the preceding action $a_{t-1}$ and the environment. The same principle applies to actions: the action dimensions of $a_t$ depend only on previous states and actions and are conditionally independent of each other. They do not exhibit a clear causal order. For instance, in a 3D coordinate $(x,y,z)$, it is not evident whether $x$ depends on $y$ or vice versa. Regarding the language prompt, as it is provided by the user, the model's role is to encode and understand it rather than generate it, akin to BERT \cite{DBLP:conf/naacl/DevlinCLT19}. Vanilla causal attention might impede information flow within each segment of a multimodal EIF trajectory, prohibiting $s_{1, t}$ from attending to $s_{2:M, t}$, and $s_{2, t}$ from attending to $s_{3:M, t}$, and so forth. This phenomenon also manifests in the prompt and action segments.

To address the issue, we propose an efficient Transformer self-attention mechanism for multimodal EIF trajectories, termed trajectory attention. Trajectory attention exhibits two key properties: the inter-segment connections are causal, and the intra-segment connections are bidirectional. Its corresponding attention matrix is illustrated in Figure~\ref{fig:matrix_comparison}. Following the convention of Transformer attention matrices, we designate the column index (top) as the source of self-attention and the row index (left) as the destination. Consequently, the causal attention matrix has all its lower triangular entries, $(i, j)$ for $i \geq j$, set to one, while the remaining entries are set to zero. Trajectory attention is achieved by unmasking the entries corresponding to $(p_{i}, p_{j})$, $(s_{i, t}, s_{j, t})$, or $(a_{i, t}, a_{j, t})$ for $i < j$. When compared with causal attention, there are 
$L(L-1)/2+T\big( M(M-1)/2 + N(N-1)/2 \big)$ additional entries joining the self-attention in every Transformer layer, which theoretically explains the effectiveness of trajectory attention. As a result, Astra is designed to process multimodal trajectories at the segment level, aligning well with the EIF setting, as it involves states and actions rather than individual tokens.

\paragraph{Action query.} Adapting to the segment-level trajectory attention mechanism, we introduce a segment-level decoding scheme based on learnable action queries. Most prior VLAs generate action dimensions autoregressively, where each action dimension depends on its preceding token \cite{DBLP:journals/corr/abs-2406-09246, DBLP:journals/corr/abs-2307-15818}. However, this approach is suboptimal because the embedding of the preceding token is highly dependent on its input and may lack the most relevant information for the action dimension. For instance, when generating $a_{1, t}$, its preceding token is $s_{M, t}$. Although the embedding of $s_{M, t}$ can aggregate information from the past trajectory through self-attention, it remains significantly influenced by its corresponding input image due to various mechanisms in Transformers, such as residual connections. Consequently, it may fail to encapsulate sufficient information necessary for accurately predicting $a_{1, t}$. 

To overcome this limitation, we adopt learnable action queries $q_{1:N}$ for individual action dimensions $a_{1:N}$, inspired by DETR \cite{DBLP:journals/mta/ChenLZT24, DBLP:conf/eccv/CarionMSUKZ20}. Each action query $q_{i}$ is dedicated to one action dimension $a_{i}$ and is shared across all timesteps: $q_{i, t=1} = q_{i, t=2} = \dots =  q_{i, t=T}$ for $i\in \{1\dots N\}$. We argue that this approach can find more relevant information for each action dimension because the action query $q_{i}$ can exclusively attend to information pertinent to $a_{i, t}$. Since action queries have no associated input token, their embeddings fully retain action dimension information. Moreover, distinct from autoregressive generation, the action queries are independent of each other and can therefore generate all dimensions of an action segment in parallel. Consequently, the decoding procedure operates at the segment level. As the action queries are solely used for information extraction and do not hold any trajectory information, they are masked out from the attention matrix, ensuring that other tokens cannot see them through the self-attention mechanism.


\paragraph{Astra.} Combining trajectory attention and action queries, we introduce a novel Transformer architecture named Astra for imitation learning. The training process optimizes the standard behavior cloning objective on offline expert trajectories:
\begin{equation}
\begin{split}
\mathcal{L}_{\text{BC}} = \min_{\theta} \sum_{t=1}^T -\log \pi_{\theta}(a_t | p, s_{\leq t}, a_{< t}).
\end{split}
\end{equation}

\input{tables/vima_bench}

\subsection{Contrastive Dynamics Learning}
Dynamics learning encourages the model to learn how the environment transitions from one state to another based on the agent's action, enabling it to make more informed decisions for EIF. Our contrastive dynamics learning (CDL) introduces minimal overhead to the model architecture, requiring only an additional classification head composed of a pooling and linear layer. As illustrated in Figure~\ref{fig:pos_neg}, we construct positive samples by augmenting the anchor trajectory using standard image augmentation and a novel action perturbation technique. Negative samples are created by mismatching states and actions from different trajectories.

Concretely, we assume that the anchor trajectory is $\tau = (p, s_{t=1}, a_{t=1}, \dots, s_{t=T}, a_{t=T})$. To construct a positive sample, $\tau^{+}$, we first apply standard computer vision data augmentation techniques to state images, such as random cropping. Additionally, we introduce a novel approach for augmenting actions. The intuition is that a slightly deviated path can still lead the agent to the desired destination, as shown in Figure~\ref{fig:action_perturb}. To achieve this, we perturb the actions by adding a small amount of random noise. By combining image augmentation and action perturbation, the positive sample is an augmented version of the anchor trajectory. 

Subsequently, we create negative trajectories that violate the correct environment dynamics. Given different trajectories from the anchor, $\tau' = (p', s'_{t=1}, a'_{t=1}, \dots, s'_{t=T}, a'_{t=T})$ and $\tau'' = (p'', s''_{t=1}, a''_{t=1}, \dots, s''_{t=T}, a''_{t=T})$, we mismatch their states and actions with those of the anchor trajectory to construct negative samples: $\tau^{-} = (p, s'_{t=1}, a_{t=1}, \dots, s_{t=T}, a''_{t=T})$. 

These strong negatives are constructed based on the following three principles discovered during the development of CDL. (1) We refrain from inserting entirely random actions or states, as these have not appeared in the dataset and can be easily identified as negatives. (2) Instead of mismatching only the original states and actions, we also use augmented ones. This prevents models from trivially identifying positive samples by detecting the presence of image augmentation or action perturbation. (3) We avoid merely shuffling states and actions along the time axis, as such negatives are also easily recognizable. 



In CDL, Astra encodes the entire multimodal trajectory into a sequence of embeddings. Due to our trajectory attention mechanism, the action tokens at the final timestep attend to the entire trajectory. Their token embeddings are then aggregated into a single trajectory embedding using a simple classification head, consisting of a pooling layer \cite{DBLP:journals/corr/abs-2311-01378} and a linear layer, as shown in Figure~\ref{fig:overview}. We denote this process as $f(\cdot)$. Finally, we employ the standard InfoNCE objective \cite{DBLP:journals/corr/abs-1807-03748} in contrastive learning to train the model to distinguish positive trajectories from negative ones:
\begin{equation}
\begin{split}
&\mathcal{L}_{\text{CDL}}(\tau, \tau^{+}, \tau^{-}) \\
= & -\log \mathbb{E} \Bigg[ \frac{ s(\tau, \tau^{+}) }{ s(\tau, \tau^{+}) + \sum_{i} s(\tau, \tau^{-}_{i}) } \Bigg],
\end{split}
\end{equation}
where $s(x,y) = \exp(f(x) \cdot f(y))$. Because \mbox{Astra} does not need to decode actions for trajectory encoding, action queries and their corresponding attention entries are not included during CDL.


During training, we incorporate CDL as an auxiliary objective alongside the primary behavior cloning objective to define the overall training loss: $\mathcal{L} = \mathcal{L}_{\text{BC}} + \alpha \mathcal{L}_{\text{CDL}}$.


%% file: tables/vima_bench.tex
\begin{table*}
\centering
    \caption{Performance comparison of success rate (\%) on the VIMA-Bench benchmark. ``Attn'' stands for attention. ``Params'' denotes the number of parameters. Gato$\ast$ modifies the original Gato model by incorporating object tokens.}
    \begin{tabular}{l | c c c | c c c c }
    \toprule
    \multicolumn{1}{c|}{} & \multicolumn{3}{c|}{\textbf{Configuration}} & \multicolumn{4}{c}{\textbf{Generalization Levels}} \\
        \textbf{Model} & \textbf{Attn Type} & \textbf{Visual Token} & \textbf{Params} & \textbf{L1} & \textbf{L2} & \textbf{L3} & \textbf{L4}\\

        \midrule
        DT & Causal & Single Image & 42.0M & 
            56.15 & 55.38 & 44.17 & 12.50 \\ 

        Gato & Causal & Image Patches & 42.0M & 
            53.08 & 50.77 & 41.67 & 15.00 \\

        Flamingo & Cross & Image Perceiver & 42.4M & 
            51.54 & 52.31 & 43.33 & 10.00 \\

        Gato$\ast$ & Causal & Object Tokens & 42.0M & 
            85.77 & 82.62 & 78.92 & 40.25 \\

        VIMA & Cross & Object Tokens & 42.4M & 
            87.69 & 86.92 & 84.17 & 47.50 \\

        \hline
        Astra (ours) & Trajectory & Object Tokens & 37.8M & 
            \textbf{97.08} & \textbf{94.62} & \textbf{86.17} & \textbf{49.50} \\

        \bottomrule
    \end{tabular}
    \label{tab:vima_bench}
\end{table*}

%% file: sections/4_experiments.tex
\section{Experiments}

\subsection{Experimental Setup}

We compare our approach with various baseline models across three different benchmarks: VIMA-Bench \cite{DBLP:journals/corr/abs-2210-03094}, ManiSkill \cite{DBLP:journals/corr/abs-2410-00425, gu2023maniskill2}, and CALVIN \cite{DBLP:journals/ral/MeesHRB22}. Each benchmark emphasizes different aspects of robot learning. VIMA-Bench investigates multimodal robot learning, where the embodied instructions are multimodal, and evaluates generalization to novel adjectives, nouns, and even meta-tasks. ManiSkill targets everyday objects with complex geometries, testing generalization to objects with unseen geometric and visual attributes. CALVIN, on the other hand, examines long-horizon manipulation tasks, assessing how well models generalize to new environments.

\input{tables/maniskill}

\input{tables/calvin}

\subsection{Implementation Details}
We introduce Astra with two model sizes. Astra (38M) is composed of 12 layers, 16 attention heads, and an embedding size of 512. Astra (198M) is composed of 10 layers, 20 attention heads, and an embedding size of 1280. As Astra is a novel architecture for Transformer backbones, it is compatible with various types of vision encoders, language encoders, and action types, as demonstrated by different configurations across the three benchmarks. Within each benchmark, the vision encoder and language encoder remain identical for Astra, DT, Gato, Flamingo, and VIMA, as we focus on comparing the core Transformer backbone. All models are trained using the AdamW optimizer \cite{DBLP:conf/iclr/LoshchilovH19}. Baselines do not incorporate CDL. We provide benchmark-specific implementation details---such as vision encoder, language encoder, and hyperparameters---in their respective sections. 

\subsection{VIMA-Bench}

VIMA-Bench \cite{DBLP:journals/corr/abs-2210-03094} focuses on multimodal robot learning, where the instructions are multimodal. We summarize the results in Table~\ref{tab:vima_bench}. It evaluates generalization capabilities across four levels: placement generalization (L1), combinatorial generalization (L2), novel object generalization (L3), and novel task generalization (L4). Each level presents increasing difficulty, with placement generalization involving only the randomization of object positions, combinatorial generalization recombining seen adjectives and nouns, novel object generalization introducing unseen adjectives or nouns, and novel task generalization incorporating entirely new meta-tasks. 

The vision encoder is ViT \cite{DBLP:conf/iclr/DosovitskiyB0WZ21} and the prompt encoder is T5 \cite{DBLP:journals/jmlr/RaffelSRLNMZLL20}. Baselines in VIMA-Bench explore different methods of encoding images, such as patch tokens and object tokens. Astra uses the best-performing object tokens as visual inputs. Discrete $SE(2)$ actions are used in this benchmark. All models are trained for 10 epochs with learning rate $= 1 \times 10^{-4}$ and weight decay $= 0.1$. Each task is evaluated using 100 trials. 

According to the VIMA-Bench paper, models with cross-attention and self-attention achieve comparable performance only when the model size exceeds 42M parameters. Therefore, we adopt this configuration for all baselines to strike a balance between model size and performance. We intentionally reduce our model size by approximately 10\%. Despite the smaller model size, Astra achieves improved performance across all four generalization levels, demonstrating its efficiency.

\input{tables/ablation}

\subsection{ManiSkill}
\label{sec:performance}

In the ManiSkill2 environment \cite{DBLP:journals/corr/abs-2410-00425, gu2023maniskill2}, we evaluate one of the most commonly utilized skills, ``pick and place'', using everyday objects with complex geometries, as detailed in Table~\ref{tab:maniskill}. Its goal is to pick up an object and place it into a container. This benchmark spans generalization levels L1 to L3 in VIMA-Bench: all items are randomly placed and the robot pose is randomly initialized, thereby including placement generalization; novel objects are also introduced as unseen tasks, facilitating both combinatorial and novel object generalization. To test these generalization capabilities, we limit the training set to 15 tasks and evaluate the models on 34 tasks. Training data includes ``2-4'' distractors. 

This benchmark consists of five types of unseen tasks. The first three types involve picked objects with unseen colors, sizes, and shapes. For example, the apple is part of the training data, while the bowl, with its novel shape, is not. The fourth type introduces unseen containers. The fifth type composes all of the first four types. A distractor is an item that is neither the picked object nor the container. They are randomly sampled from a diverse pool of items. For all five types of ``unseen tasks'', we randomly sample and place 2-4 distractors. For seen objects, we explore whether the number of distractors can impact the models' performance. 

We utilize ResNet \cite{DBLP:conf/cvpr/HeZRS16} for images and T5 \cite{DBLP:journals/jmlr/RaffelSRLNMZLL20} for language prompts. Actions are discrete 6D poses. The models are trained for 5 epochs with learning rate $= 1 \times 10^{-4}$ and weight decay $= 1 \times 10^{-4}$. We conduct 50 trials for each task, and each trial is limited to 100 timesteps before a timeout. 

We experiment with the same number of Transformer layers for Astra (198M), VIMA, and Gato. RT-1 retains its original configuration with 46M parameters. Our model matches the model size of Gato while achieving superior performance. Due to the additional cross-attention layers in VIMA, its model size is significantly larger, which may have contributed to overfitting in this experiment.

\subsection{CALVIN}
The CALVIN benchmark \cite{DBLP:journals/ral/MeesHRB22} focuses on long-horizon manipulation tasks. Performance comparison results are presented in Table~\ref{tab:calvin}. During each evaluation session, the model is prompted with five random tasks. The session terminates as soon as a task fails, and the remaining tasks are not attempted. Performance is measured by the number of tasks successfully completed in a row. The benchmark provides three different experimental settings: D$\rightarrow$D, ABCD$\rightarrow$D, and ABC$\rightarrow$D, where each letter represents a distinct environment. For example, in the ABC$\rightarrow$D setting, the model is trained on data from environments A, B, and C, but evaluated in environment D. This setting thus assesses zero-shot generalization to new environments. We compare our model to baselines in this most challenging ABC$\rightarrow$D experiment. Since only 1\% of the training dataset for ABC$\rightarrow$D is annotated with language prompts, we utilize this language-annotated subset for training, further increasing the difficulty of the task. 

We use MAE-ViT vision encoder \cite{DBLP:conf/cvpr/HeCXLDG22} and CLIP language encoder \cite{DBLP:conf/icml/RadfordKHRGASAM21}, following GR-1 \cite{DBLP:conf/iclr/WuJCCXLLLK24}. All models use continuous 6D pose actions and are trained from scratch for 20 epochs with learning rate $= 9 \times 10^{-4}$ and weight decay $= 1 \times 10^{-4}$. 

Since the rollout of a trajectory terminates as soon as any of the five tasks fail, completing all five tasks is highly challenging. Our model can complete longer task sequences, highlighting its effectiveness in generalizing to new environments.

\begin{figure*}
    \centering
     \begin{subfigure}[b]{0.49\textwidth}
         \centering
            \includegraphics[width=\columnwidth, trim={0cm 0.2cm 0cm 0.2cm},clip]{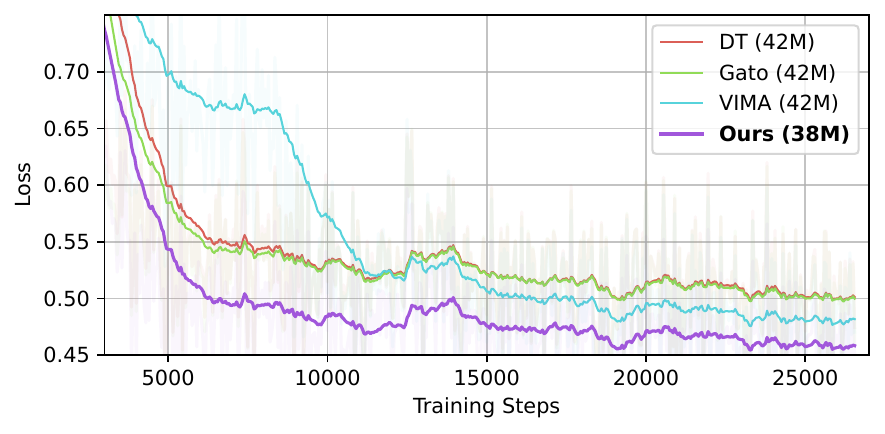}
     \end{subfigure}
     \begin{subfigure}[b]{0.49\textwidth}
         \centering
            \includegraphics[width=\columnwidth, trim={0cm 0.2cm 0cm 0.2cm},clip]{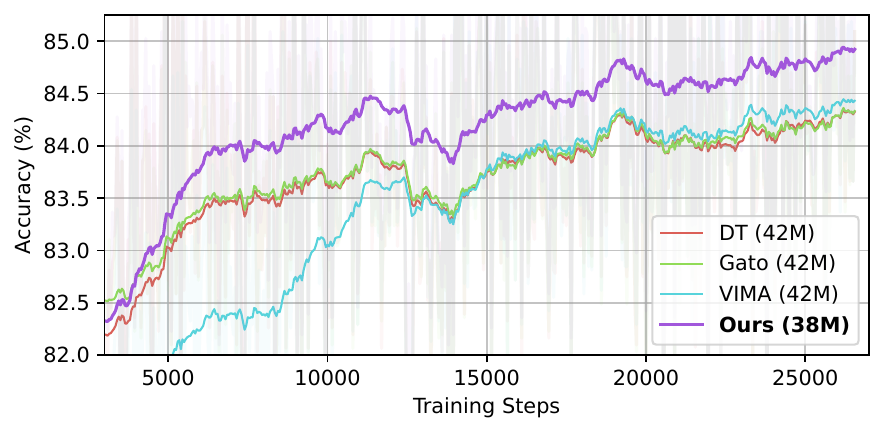}
     \end{subfigure}
    \caption{Loss and accuracy curves during training on VIMA-Bench.}
    \label{fig:curves}
\end{figure*}

\begin{figure*}
    \centering
        \includegraphics[width=\textwidth, trim={0.0cm 24.2cm 4.2cm 0.75cm},clip]{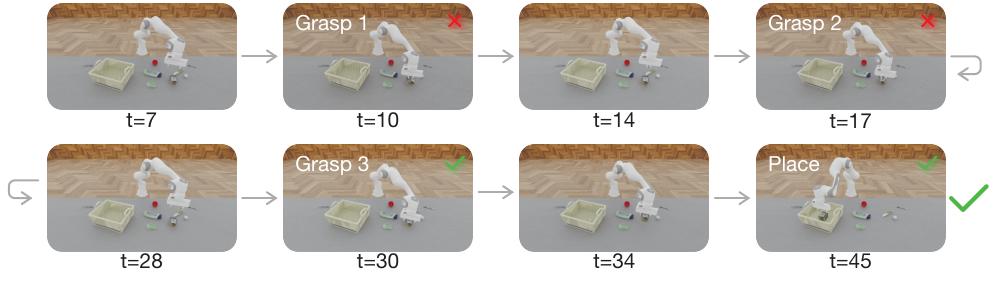}
        \caption{An example of instantaneous regrasp. The task is to ``pick blue tea box and place into clear box.'' Three grasp attempts were completed within only 30 steps, a capability not observed in the baseline models.}
        \label{fig:regrasp}
\end{figure*}

\subsection{Ablation Study}
\label{sec:ablation}

We analyze the effects of the proposed components of Astra, including trajectory attention, action queries, and contrastive dynamics learning, as shown in Table~\ref{tab:ablation}. Since our trajectory data augmentation method is only applied in conjunction with CDL, it is not utilized for ``Astra'' (rows 2-5). In ManiSkill, we provided more granular results across three difficulty levels (Appendix~\ref{appendix:difficulty_levels}). 

CDL proves effective in enhancing imitation learning, particularly for the first three generalization levels, as it enables the model to better learn environment dynamics. The lower performance on L4 may be attributed to CDL's training data, which includes only seen nouns and adjectives, thereby enhancing the performance on seen meta-tasks at the expense of generalizability to novel ones. The removal of trajectory attention results in a noticeable decrease in success rates across all levels, underscoring its crucial role in processing segmented multimodal trajectories. Similarly, the absence of action queries leads to reduced success rates, highlighting its importance in enhancing information extraction for action generation. When both components are removed, the model reverts to a typical token-level autoregressive model, akin to Gato.

\subsection{Qualitative Analysis}
\label{sec:quality}

We present the loss and accuracy curves for the models on VIMA-Bench in Figure~\ref{fig:curves}. Despite having a model size approximately 10\% smaller, our model exhibits a faster convergence rate. The substantially lower loss and higher accuracy account for Astra's superior performance. Specifically, trajectory attention facilitates improved information flow within each segment, while action queries more effectively extract information relevant to individual action dimensions. The Flamingo baseline is excluded from the figure due to its significantly worse performance. 

In ManiSkill, we identified a crucial distinction between Astra's capabilities and those of the baselines: Astra masters ``instantaneous regrasp''. Figure~\ref{fig:regrasp} illustrates the most challenging task in ManiSkill. Baseline models often struggle to recognize failed grasps. In contrast, our Astra model promptly detects a failed grasp and repeatedly attempts to grasp the object until successful. A more detailed description is provided in Appendix~\ref{appendix:record_explanation}.

%% file: tables/maniskill.tex
\begin{table*}
\centering
    \caption{Performance comparison of success rate (\%) on the ManiSkill benchmark. ``Traj.'' is an abbreviation for trajectory attention. ``Cont.'' stands for container. ``0'', ``2-4'', ``6-8'' indicate the number of distractor objects.}
    \begin{tabular}{l | c c | c c c c c | c c c }
    \toprule
    \multicolumn{1}{c|}{} & \multicolumn{2}{c|}{\textbf{Configuration}} & \multicolumn{5}{c|}{\textbf{Unseen Tasks}} & \multicolumn{3}{c}{\textbf{Seen Tasks}} \\

        \textbf{Model} & \textbf{Attn} & \textbf{Params} & \textbf{Color} & \textbf{Size} & \textbf{Shape} & \textbf{Cont.} & \textbf{All} & \textbf{0} & \textbf{2-4} & \textbf{6-8} \\

        \midrule
        RT-1 & Causal & 46M & 
            27.03 & 6.36 & 20.30 & 0.79 & 1.27 & 
            61.09 & 39.17 & 23.40 \\

        VIMA & Cross & 525M & 
            26.00 & 26.00 & 17.20 & 30.75 & 19.33 &
            47.93 & 41.47 & 36.33 \\

        Gato & Causal & 198M & 
            46.00 & 74.00 & 42.00 & 44.40 & 40.00 &
            76.40 & 73.33 & 62.67 \\

        \hline
        Astra (ours) & Traj. & 198M & 
            \textbf{72.00} & \textbf{91.00} & \textbf{52.40} & \textbf{63.43} & \textbf{70.67} &
            \textbf{90.93} & \textbf{90.53} & \textbf{79.07} \\

        \bottomrule
    \end{tabular}
    \label{tab:maniskill}
\end{table*}

%% file: tables/calvin.tex
\begin{table*}
\centering
    \caption{Performance comparison on the CALVIN benchmark under the most challenging ABC$\rightarrow$D setting.
    }
    \begin{tabular}{l | c c | c c c c c | c}
    \toprule
    \multicolumn{1}{c|}{} & \multicolumn{2}{c|}{\textbf{Configuration}} & \multicolumn{5}{c|}{\textbf{Tasks completed in a row (\%)}} & \multicolumn{1}{c}{} \\

        \textbf{Model} & \textbf{Attn Type} & \textbf{Params}  & \textbf{1} & \textbf{2} & \textbf{3} & \textbf{4} & \textbf{5} & \textbf{Avg. Len.} \\

        \midrule

        MCIL & \NA & 63.6M & 
            30.4 & 1.3 & 0.2 & 0.0 & 0.0 & 0.31 \\

        OpenVLA & Causal & 7B &
            32.4 & 4.2 & 1.4 & 0.5 & 0.3 & 0.39 \\

        RT-1 & Causal & 46M &
            53.3 & 22.2 & 9.4 & 3.8 & 1.3 & 0.90 \\

        MDT & Cross & 75.1M &
            61.7 & 41.6 & 23.8 & 14.7 & 8.7 & 1.54 \\

        RoboFlamingo & Cross & 1B &
            82.4 & 61.9 & 46.6 & 33.1 & 23.5 & 2.48 \\

        GR-1 & Causal & 21.3M &
            85.4 & 71.2 & 59.6 & 49.7 & 40.1 & 3.06 \\

        \hline

        VIMA & Cross & 42.4M  & 
            64.1 & 47.6 & 34.6 & 29.2 & 23.7 & 1.99 \\

        Flamingo & Cross & 42.4M & 
            69.9 & 53.2 & 39.1 & 31.2 & 24.6 & 2.18 \\

        DT & Causal & 44.1M &
            74.1 & 56.1 & 42.0 & 32.2 & 25.6 & 2.30 \\

        Gato & Causal & 44.1M &
            77.3 & 57.0 & 44.4 & 33.4 & 26.1 & 2.38 \\

        \hline
        Astra (ours) & Trajectory & 37.8M &
            \textbf{89.7} & \textbf{79.2} & \textbf{65.8} & \textbf{52.4} & \textbf{42.3} & \textbf{3.29} \\

        \bottomrule
    \end{tabular}
    \label{tab:calvin}
\end{table*}

%% file: tables/ablation.tex
\begin{table*}
    \centering
    \caption{Ablation study of the components in Astra. In ManiSkill, we compare seen tasks with 2-4 distractors.}
    \begin{tabular}{l | c c c c | c c c}
    \toprule
        \multicolumn{1}{c|}{} & \multicolumn{4}{c|}{\textbf{VIMA-Bench}} & \multicolumn{3}{c}{\textbf{ManiSkill}} \\

        \textbf{Configuration} & \textbf{L1} & \textbf{L2} & \textbf{L3} & \textbf{L4} & \textbf{Easy} & \textbf{Medium} & \textbf{Hard} \\
        \midrule
            Astra w/ CDL &
                \textbf{97.08} & \textbf{94.62} & \textbf{86.17} & 49.50 &
                \textbf{93.83} & \textbf{89.71} & \textbf{83.50} \\

            \hline
            Astra & 
                94.69 & 92.15 & 85.83 & \textbf{50.00} & 
                91.33 & 88.57 & 75.00 \\

            \hline
             \hspace{0.25cm} w/o Trajectory Attention & 
                91.23 & 89.31 & 84.42 & 47.50 & 
                86.33 & 83.71 & 76.00 \\

            \hline
             \hspace{0.25cm} w/o Action Query & 
                89.08 & 84.85 & 82.58 & 45.25 & 
                86.00 & 80.86 & 69.00 \\

            \hline
             \hspace{0.25cm} w/o Both & 
                85.77 & 82.62 & 78.92 & 40.25 &
                72.33 & 76.00 & 67.00 \\
    \bottomrule
    \end{tabular}
    \label{tab:ablation}
\end{table*}

%% file: sections/appendix.tex
\section{Appendix}

\subsection{Notation}
In this paper, we use the following notation:

\input{tables/notation}

\subsection{Trajectory Attention Matrix}
\begin{figure}[t]
    \centering
     \begin{subfigure}[t]{1.0\columnwidth}
         \centering
            \includegraphics[width=\columnwidth, trim={0.625in 2.5in 0in 0.625in},clip]{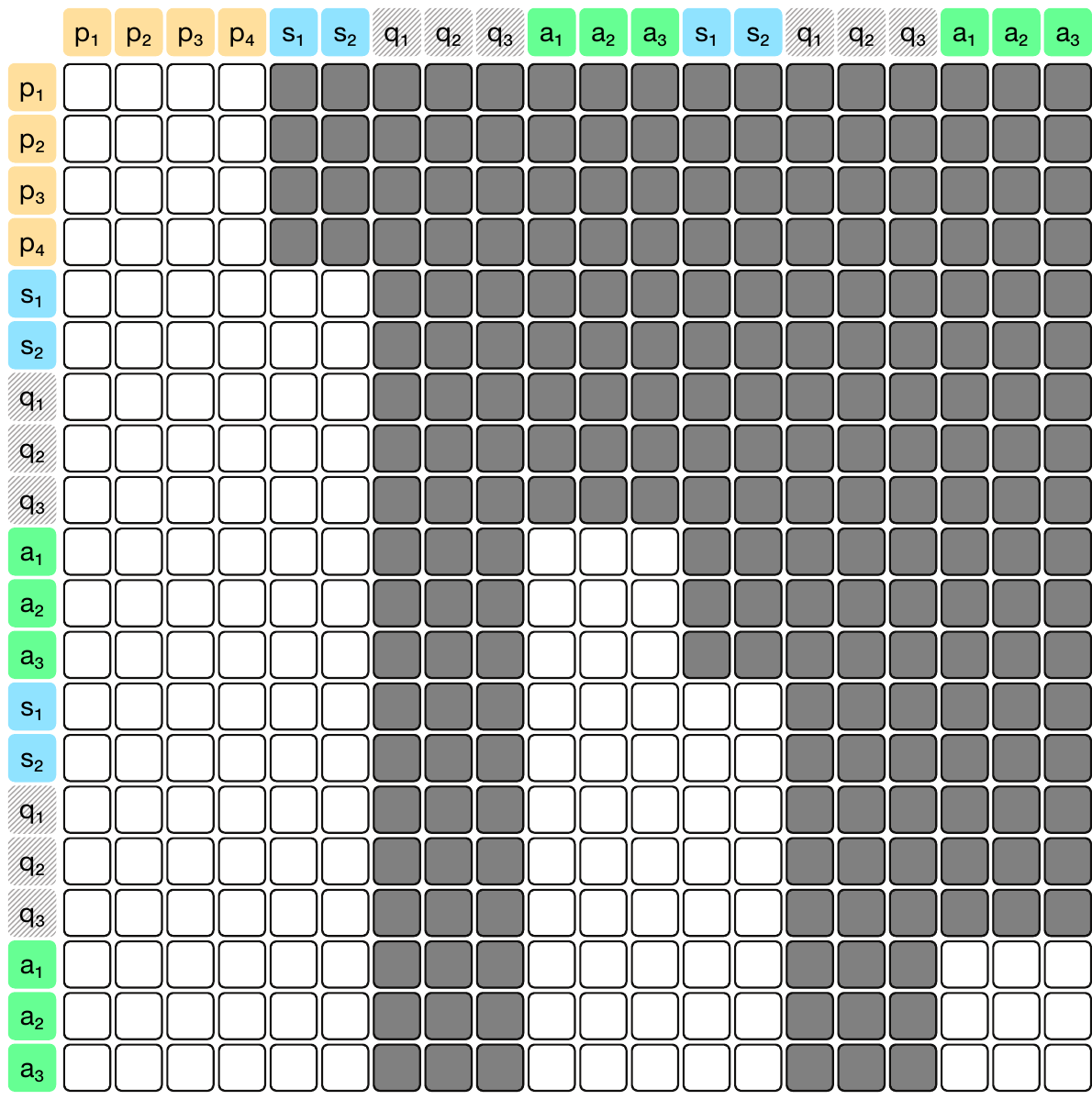}
            \caption{Trajectory attention matrix for decoding.}
            \vspace{2pt}
     \end{subfigure}
     \hfill
     \begin{subfigure}[t]{0.714\columnwidth}
         \centering
            \includegraphics[width=\columnwidth, trim={0.625in 4.75in 2.25in 0.625in},clip]{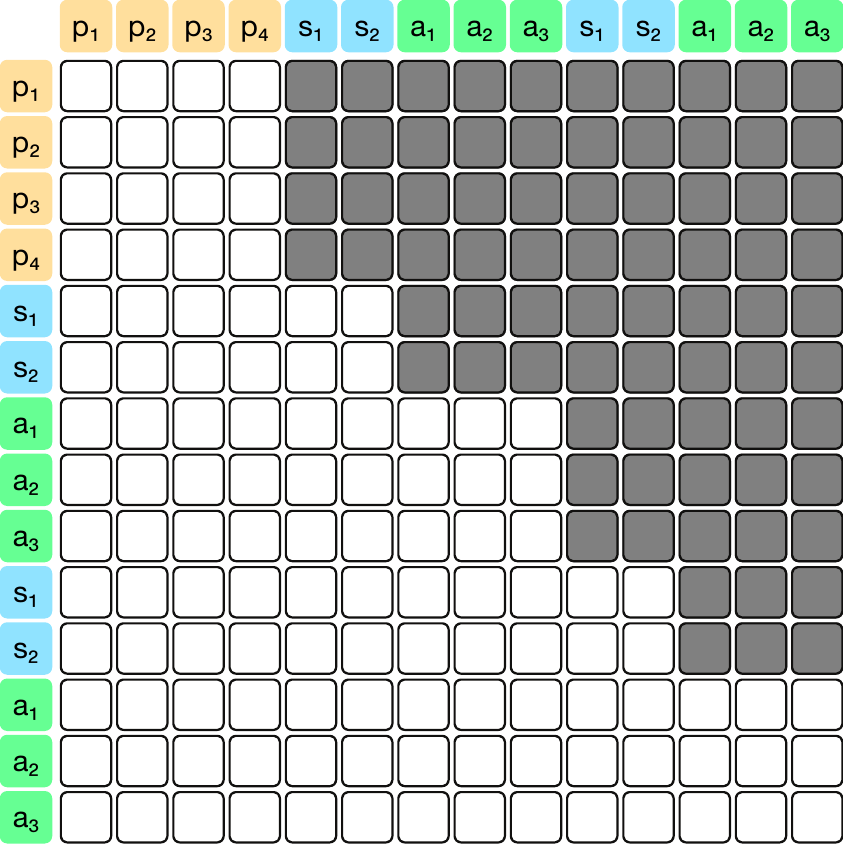}
            \caption{Trajectory attention matrix for encoding.}
     \end{subfigure}
    \caption{The attention matrices of trajectory attention. Input tokens are at the top, and output tokens are on the left. Dark boxes represent masked entries in the attention matrices. The decoding attention matrix is utilized in imitation learning, whereas the encoding attention matrix is employed in contrastive dynamics learning. }
     \label{fig:attn_matrix}
\end{figure}

The trajectory attention is implemented with its corresponding attention mask, as illustrated in Figure~\ref{fig:attn_matrix}. Output tokens on the left attend to input tokens at the top. For example, in the first row, the token $p_1$ can attend to tokens $(p_1, p_2, p_3, p_4)$ but not to any other subsequent tokens. Consequently, tokens starting from $s_1$ onward are masked out. During action generation in imitation learning, action queries can attend to all tokens up to the current state, while no other tokens can attend to action queries. Therefore, action query tokens are completely masked from the input.

Although we can use the decoding attention matrix for both decoding and encoding, utilizing the encoding attention matrix can avoid computational overhead for processing action queries. Regardless of whether the decoding or encoding matrix is employed, the resulting embeddings for the encoded trajectory remain identical in contrastive dynamics learning.

\begin{figure*}
    \centering
     \begin{subfigure}[b]{\columnwidth}
         \centering
            \includegraphics[width=\textwidth, trim={0cm 1cm 0cm 0cm},clip]{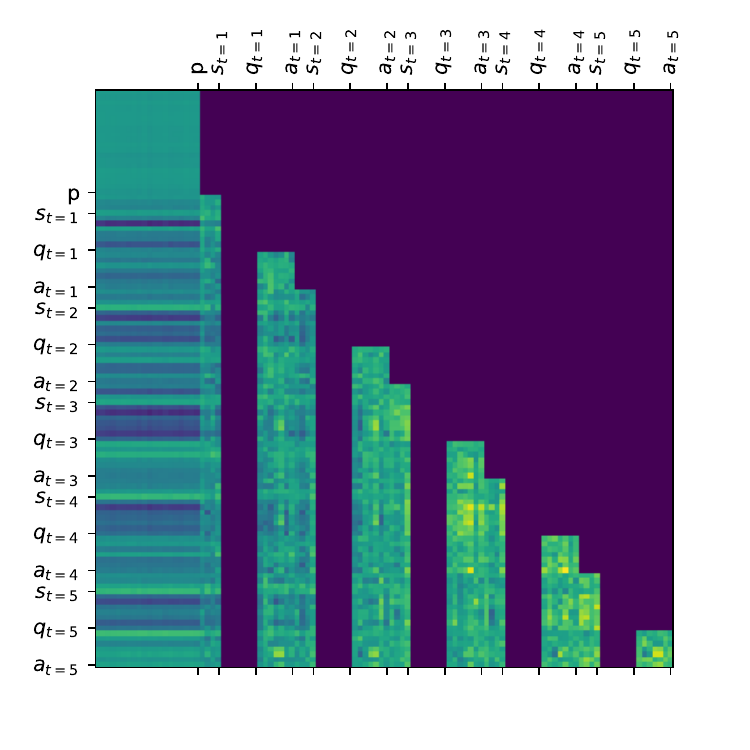}
     \end{subfigure}
     \hfill
     \begin{subfigure}[b]{\columnwidth}
         \centering
            \includegraphics[width=\textwidth, trim={0cm 1cm 0cm 0cm},clip]{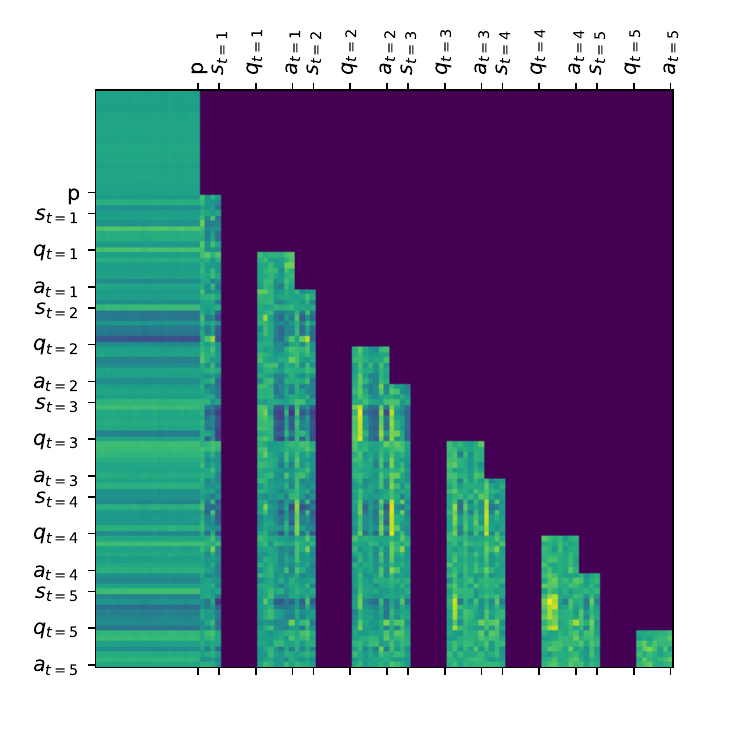}
     \end{subfigure}
    \caption{Visualization of two trained trajectory attention matrices over five timesteps. Brighter cells indicate higher attention weights. Tick labels are displayed for the last token of each segment.}
    \label{fig:visualization}
\end{figure*}

\subsection{Trajectory Attention Visualization}

We present a visualization of the attention matrices from the top layer of Astra, depicted in Figure~\ref{fig:visualization}. Prompt tokens exhibit similar attention values, while state and action tokens from more recent timesteps receive higher attention weights compared to those from earlier in the sequence. This observation aligns with the expectation that the latest timestep provides the most informative context for generating the next action. Moreover, several attention weights above the main diagonal are strongly activated, suggesting that the additional attention connections enabled by our trajectory attention mechanism are beneficial. In the second matrix, a clear distinction emerges between the attention weights for the output state tokens and query tokens. This distinction highlights that action queries extract information differently from state tokens, elucidating their role in improving action generation.

\subsection{Integration with Large VLAs}
Due to the trajectory attention mechanism and action queries in Astra, large VLA models that rely on autoregressive modeling cannot directly adopt this architecture without significant retraining. However, there are alternative methods to leverage the Astra architecture in large VLAs. For instance, it can be incorporated as an action prediction head by attaching it to the top of pretrained VLMs. The integration of Astra with large VLA models presents a promising direction for future research.


\subsection{ManiSkill Generalization Levels}
\label{appendix:general_levels}

For Astra, Gato, and VIMA in Table~\ref{tab:maniskill}, unseen shape proves to be the most challenging generalization level of ManiSkill, followed by unseen containers. VIMA also exhibits volatility when dealing with small objects from the ``Size'' level. RT-1, in particular, struggles with identifying the container when an unseen container is introduced. It is important to note that not all seen target objects (the object to be picked) have a generalized version. For example, a strawberry is a seen target object that is difficult to grasp, but there is no oversized strawberry for the ``Shape'' level. Consequently, models may achieve a higher success rate on some generalization levels than on seen tasks. Furthermore, since unseen color and size are part of the mixture, the success rate in ``All'' is not as low as in ``Shape'' or ``Container''. The introduction of more distractors in the scene increases the likelihood of collisions and causes additional difficulty in grasping the objects. However, this negative effect is not severe enough to considerably degrade the performance.


\subsection{ManiSkill Difficulty Levels}
\label{appendix:difficulty_levels}

The ablation study in the ManiSkill environment is reported by difficulty level, as shown in Table~\ref{tab:ablation}. In simple terms, easy tasks involve spherical, regular-sized target objects, such as a baseball. The medium difficulty level includes elongated or small target objects, such as a banana or a strawberry. Hard tasks encompass oversized, non-spherical, or thin objects, such as a tea box or knife. 

Easy tasks include spherical, regular-sized objects. Round objects are easier to pick because the robot arm can close the gripper in any direction. Size also has a significant impact on the success rate because oversized objects require more precise grasp poses. If a grasp is not precise, the two fingers of the gripper may collide with the object and not be able to reach down on the object. On the other hand, small objects can increase the difficulty because the gripper might miss them if the grasp is slightly off. An elongated object, such as a remote controller or a banana, must be picked up ``across'' the object rather than ``along'' the object. Consequently, we define the medium difficulty level as the objects that are too big, too small, or elongated. Hard tasks involve non-spherical, oversized, or thin objects, such as a tea box and a knife. A tea box, being both non-spherical and oversized, requires the robot arm to grasp it precisely parallel to its sides rather than diagonally. A knife can be hard since it is very thin and close to the desk. During grasping, the gripper may collide with the desk, further increasing the difficulty.

\subsection{Instantaneous Regrasp}
\label{appendix:record_explanation}

We provide a more detailed explanation for the example in Figure~\ref{fig:regrasp}. Only three of the four grasp attempts are shown, with the third one omitted due to page limitations. Therefore, the grasp labeled as ``Grasp 3'' in the figure is, in fact, the fourth grasp attempt. The first grasp was unsuccessful because one finger of the gripper collided with the blue tea box, and the grasp slipped. Subsequently, Astra swiftly initiated two additional grasps; however, the gripper closed too early, resulting in collisions with the tea box again. Shortly after the second and third failures, the gripper's fingers successfully reached down to opposite sides of the blue tea box, completing the fourth regrasp. Remarkably, all four grasp attempts were executed within a mere 30 timesteps, a feat not achieved by any of the baseline methods.

\begin{figure}[t]
    \centering
        \includegraphics[width=1\columnwidth, trim={0cm 5.9in 0cm 0.4in},clip]{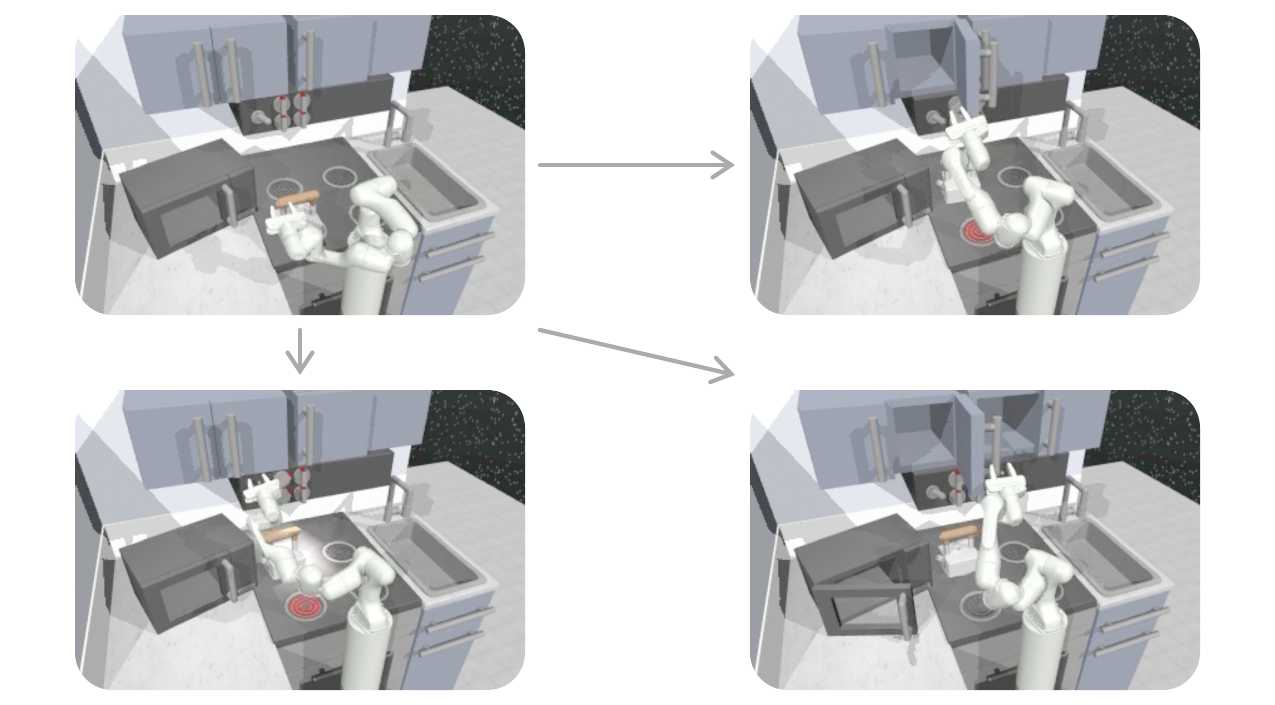}
        \caption{Astra in Franka Kitchen. The model completes four random tasks. The top left image shows the initial state and the other three images are three different final states.}
        \label{fig:kitchen_example}
\end{figure}

\begin{figure}[t]
    \centering
        \includegraphics[width=1\columnwidth, trim={0cm 2.4in 0cm 6.4in},clip]{figures_app/more_record.pdf}
        \caption{Astra in Push-T. The gray T is the object, and the green T is its target position. The model controls the blue dot to push the T-shaped object towards the target position. The red cross is the cursor. The images on the left and right are two different initial states, and the middle image is the final state where the object perfectly overlaps with the target position.}
        \label{fig:pusht_example}
\end{figure}

\input{tables/franka_kitchen}

\input{tables/push_T}

\subsection{Additional Related Work}
\label{appendix:additional_related}

VLAs \cite{DBLP:journals/corr/abs-2405-14093, DBLP:journals/corr/abs-2405-14093} have recently emerged as a new type of multimodal model for EIF tasks in the field of embodied AI. MOO \cite{DBLP:journals/corr/abs-2303-00905} introduced multi-modal prompt capability to RT-1, while Q-Transformer \cite{DBLP:journals/corr/abs-2309-10150} adapted RT-1 to the Q-learning setting. RoboFlamingo \cite{DBLP:journals/corr/abs-2311-01378} constructed a VLA based on the existing Flamingo VLM \cite{DBLP:conf/nips/AlayracDLMBHLMM22, DBLP:journals/corr/abs-2308-01390}. ACT \cite{DBLP:conf/rss/ZhaoKLF23} adopts the DETR framework for robotics tasks but utilizes fixed position embeddings at the timestep level. VLAs can also be integrated with high-level planners to address long-horizon robotics tasks, as demonstrated by SayCan \cite{DBLP:conf/corl/IchterBCFHHHIIJ22}, PaLM-E \cite{DBLP:conf/icml/DriessXSLCIWTVY23}, and ChatGPT for Robotics \cite{DBLP:journals/corr/abs-2306-17582}. Similar to other multimodal models, the success of VLAs is predicated on a foundation of numerous prior unimodal models and a variety of deep learning techniques \cite{DBLP:conf/aaai/MaSHLZK23, DBLP:conf/naacl/MaCLSZK24, DBLP:conf/nips/SongZK23a, DBLP:conf/kdd/ZhangZS00MKK24, DBLP:conf/nips/SongZK23, DBLP:conf/ijcnn/MaYLRD21}.

In addition to the primary learning objective, auxiliary or pretraining objectives have proven useful in further enhancing model performance. The success of masked language modeling, as initially proposed in BERT \cite{DBLP:conf/naacl/DevlinCLT19}, has prompted the adoption of similar objectives in various domains. In computer vision models and VLMs, representative works like MAE \cite{DBLP:conf/cvpr/HeCXLDG22} and ViLBERT \cite{DBLP:conf/nips/LuBPL19} have employed comparable strategies. VLAs have also utilized masked modeling objectives for their vision encoders, such as MVP \cite{DBLP:conf/corl/RadosavovicXJAM22}, Voltron \cite{DBLP:conf/rss/KaramchetiNCKFS23}, GR-1 \cite{DBLP:conf/iclr/WuJCCXLLLK24}. While these approaches have proven beneficial for the vision encoder, they often overlook the crucial alignment between different modalities. 

Dynamics learning has long been recognized as a powerful technique for improving the performance of robot learning models. Dreamer \cite{DBLP:conf/iclr/HafnerLB020} was a pioneering work in this domain, inspiring several follow-up methods, including Iso-Dream \cite{DBLP:conf/nips/Pan0WY22}, TWM \cite{DBLP:conf/iclr/RobineHUH23}, and IRIS \cite{DBLP:conf/iclr/MicheliAF23}.



\subsection{Additional Baselines}

The ``Diffusion'' baseline in Appendix~\ref{appendix:more_record} is based on a causal Transformer backbone and is trained with modified training objectives: DDPM \cite{DBLP:conf/nips/HoJA20} for continuous actions and D3PM \cite{DBLP:conf/nips/AustinJHTB21} for discrete actions. The training losses are defined as follows:
\begin{equation}
\begin{split}
\mathcal{L}_{\text{DDPM}} &= \operatorname{MSE}\left(\varepsilon^k, \varepsilon_\theta(\mathbf{x}^0+\varepsilon^k, k) \right), \\
\mathcal{L}_{\text{D3PM}} &= \operatorname{CE}\left(\varepsilon^k, \varepsilon_\theta(\mathbf{x}^0+\varepsilon^k, k) \right),
\end{split}
\end{equation}
where $\mathbf{x}^0$ is the original action and $\varepsilon^k$ is the noise of the $k$-th iteration; $\varepsilon_\theta$ is the Transformer backbone.

We have also experimented with $\pi_0$ \cite{DBLP:journals/corr/abs-2410-24164} on the CALVIN benchmark. However, its performance was notably poor, and as a result, we decided not to include it in Table~\ref{tab:calvin}.

\subsection{Additional Benchmarks}
\label{appendix:more_record}


\paragraph{Franka Kitchen.} Franka Kitchen \cite{DBLP:conf/corl/0004KLLH19} includes five skills that span seven specific tasks within the scene. The ``turn knob'' skill involves turning the oven knob to activate either the top or bottom burner. ``Toggle switch'' involves turning on the light switch. ``Slide door open'' requires opening the slide cabinet, while ``swing door open'' involves opening either the left hinge cabinet or the microwave door by the door handle. The ``lift by handle'' skill entails moving the kettle by its handle. Figure~\ref{fig:kitchen_example} provides examples of executions by Astra in Franka Kitchen.

Performance is measured by the completion of multi-stage tasks, as summarized in Table~\ref{tab:franka_kitchen}. Results are averaged over 50 runs. In each run, the models are required to complete four random tasks within 280 steps. $p\text{i}$ means the model has completed i tasks and thus reached the i-th stage. Since the scale of Franka Kitchen is relatively small, we compare the models using the Astra (43M) configuration. Astra (43.3M parameters) consists of 6 layers, 12 attention heads, and an embedding size of 768. We also compare the performance of models using continuous and discrete actions in this environment.

\paragraph{Push-T.} In Push-T \cite{DBLP:conf/corl/FlorenceLZRWDWL21}, the models need to push a T-shaped object until it aligns perfectly with the target position. This task requires precise control, as performance is measured by the overlapping area, with perfect alignment equating to a score of 1.0. Several push-T examples by Astra are included in Figure~\ref{fig:pusht_example}.

The performance of the models is compared in Table~\ref{tab:push_T}. Results are averaged over 30 trials. The maximum number of steps the models can take in each trial is 200, so they need to push the object precisely while maintaining adequate speed. Because this is a 2D task, we determined that the configuration of Astra (19M) is adequate for most models, except for the diffusion-based model, which utilizes the configuration of Astra (43M). Astra (19.4M parameters) consists of 6 layers, 8 attention heads, and an embedding size of 512. As the cross-attention layers in VIMA result in a model size of 50.8M parameters---exceeding the size of Astra (43M)---we instead employ three Transformer blocks for VIMA. We found that models fail to learn effective policies using continuous actions; therefore, we only report results of discrete actions.

\subsection{Scalability}
Astra is an efficient Transformer architecture designed for EIF tasks. Consequently, it also inherits the scalability of Transformers. Depending on the complexity of the EIF benchmarks, we select the most efficient configuration with sufficient capacity to address the demands of each benchmark. In this paper, we explore four configurations of Astra, with model sizes ranging from 19M to 198M, tailored to benchmarks of varying levels of difficulty:
\begin{itemize}
    \item Astra (19M): Push-T (Table~\ref{tab:push_T})
    \item Astra (38M): VIMA-Bench (Table~\ref{tab:vima_bench}) and CALVIN (Table~\ref{tab:calvin})
    \item Astra (43M): Franka Kitchen (Table~\ref{tab:franka_kitchen})
    \item Astra (198M): ManiSkill (Table~\ref{tab:maniskill})
\end{itemize}
With the help of CDL, the performance of Astra (38M) and Astra (198M) can be further improved, as shown in Table~\ref{tab:ablation}. This suggests that CDL can serve as an effective large-scale pretraining method for VLAs.

\subsection{Additional Qualitative Analysis}
\begin{figure}[ht!]
    \centering
    \begin{subfigure}[b]{1.0\columnwidth}
         \centering
            \includegraphics[width=1.0\textwidth, trim={0.25in 6.25in 1.75in 0.25in},clip]{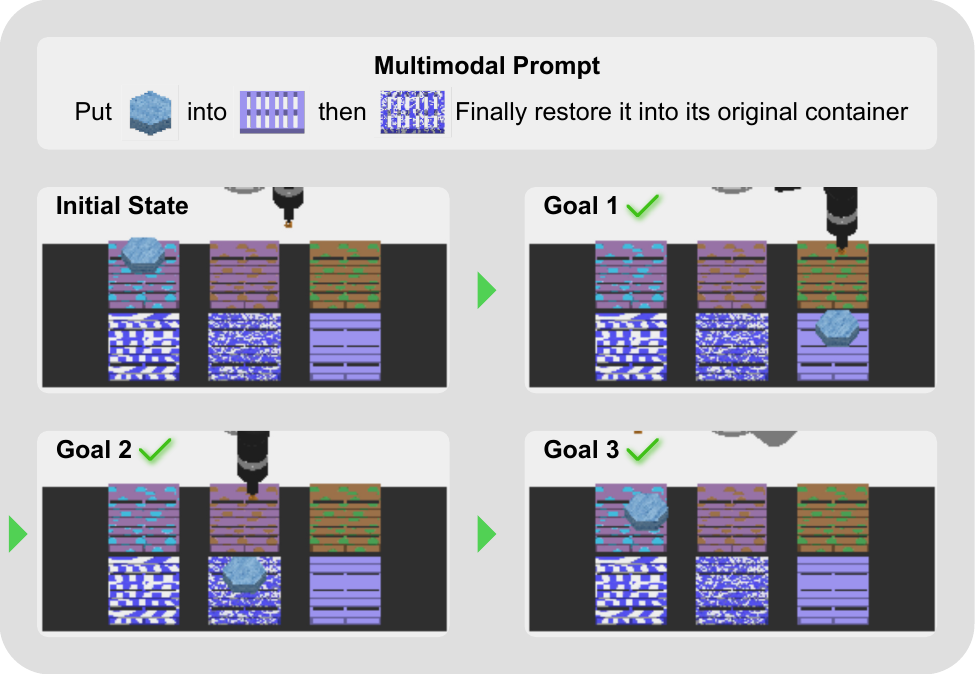}
            \caption{Astra with CDL.}
         \vspace{10pt}
     \end{subfigure}
     \begin{subfigure}[b]{1.0\columnwidth}
         \centering
            \includegraphics[width=1.0\textwidth, trim={0.25in 2.625in 1.75in 5.5in},clip]{figures_app/vb_cdl.pdf}
            \caption{Astra without CDL.}
     \end{subfigure}
    \caption{Comparison between Astra with and without contrastive dynamics learning.}
    \label{fig:vb_cdl}
\end{figure}

We provide two examples in Figure~\ref{fig:vb_cdl} that compare Astra with CDL to Astra without CDL, aiming to qualitatively demonstrate the effectiveness of contrastive dynamics learning. The failure case of Astra without CDL illustrates how CDL can contribute to improved generalization. In this comparison, CDL enables Astra to more effectively handle novel objects.

\subsection{Additional Examples}
\begin{figure}[ht!]
    \centering
        \includegraphics[width=\columnwidth, trim={0.25in 6.875in 1.75in 0.25in},clip]{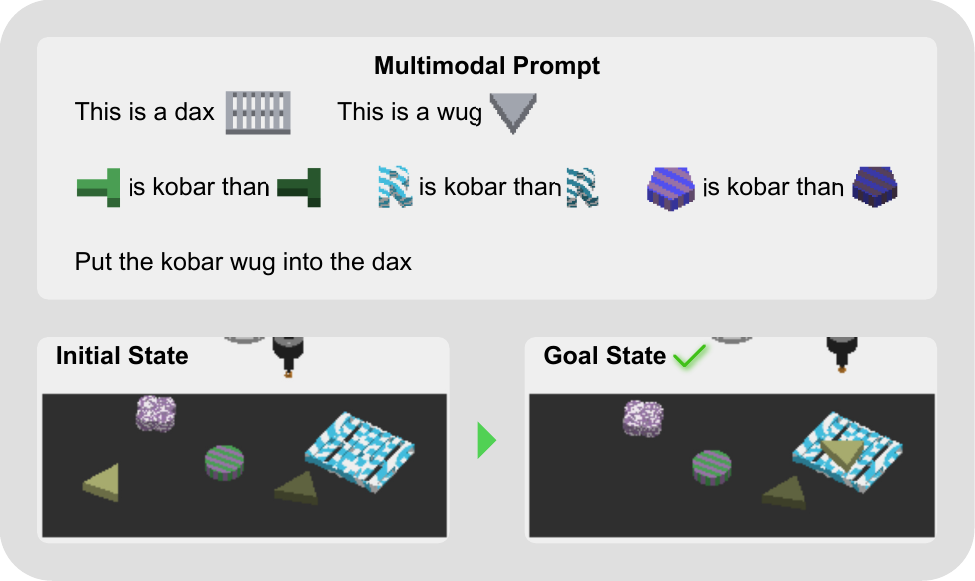}
        \caption{An example of an L4 generalization level task (novel task generalization of ``novel adjective and noun'') in VIMA-Bench.}
        \label{fig:vb_ex1}
\end{figure}

\begin{figure}[ht!]
    \centering
        \includegraphics[width=\columnwidth, trim={0.25in 6.625in 1.75in 0.25in},clip]{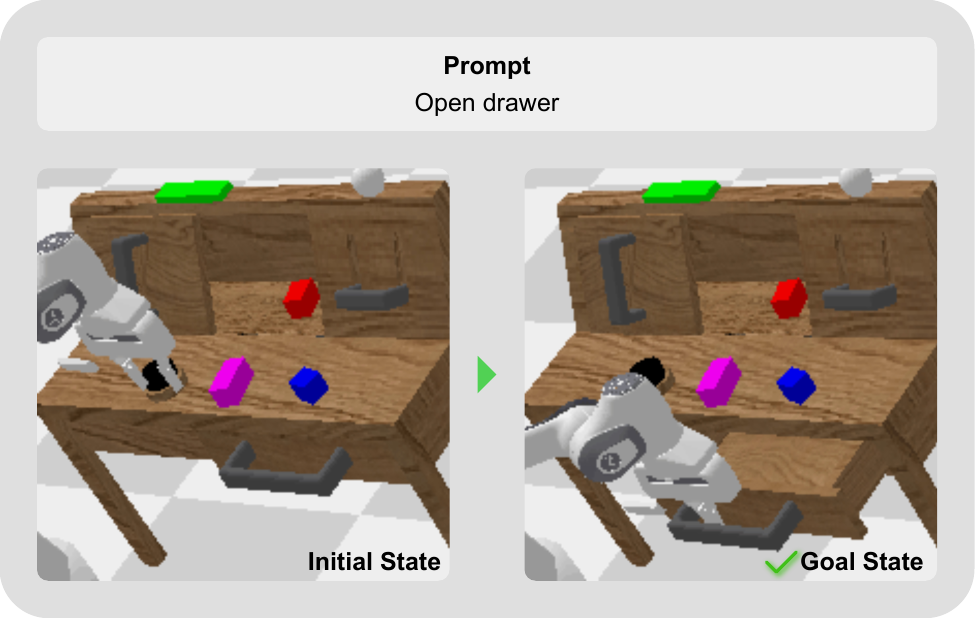}
        \caption{An example of a task in CALVIN.}
        \label{fig:calvin_ex1}
\end{figure}

\begin{figure}[ht!]
    \centering
        \includegraphics[width=\columnwidth, trim={0.25in 0.625in 1.75in 0.25in},clip]{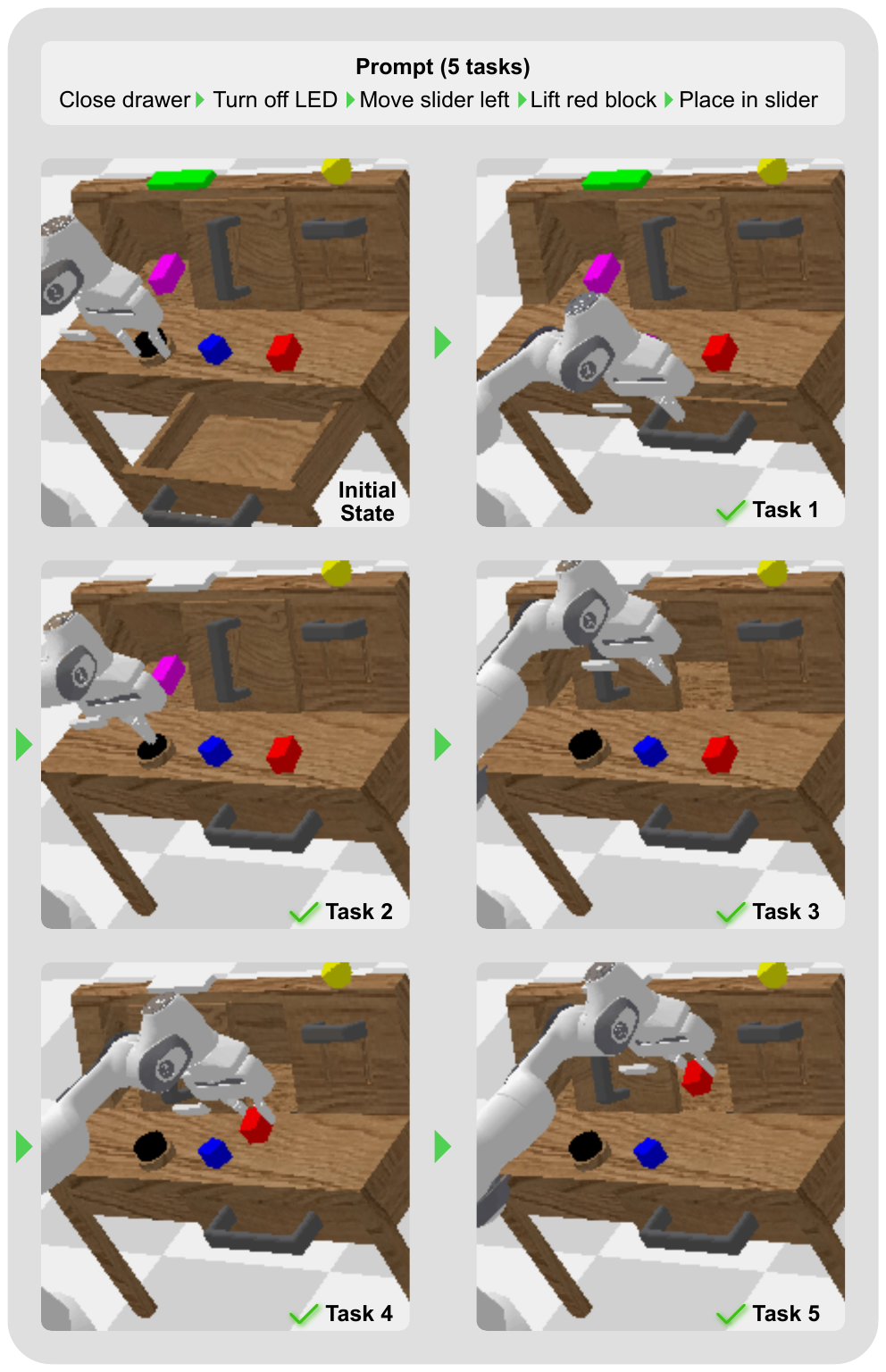}
        \caption{An example of a long-horizon trajectory in CALVIN.}
        \label{fig:calvin_ex3}
\end{figure}

We include a few execution examples of Astra in VIMA-Bench tasks in Figure~\ref{fig:vb_ex1}~\&~\ref{fig:vb_ex2} and CALVIN tasks in Figure~\ref{fig:calvin_ex1}~\&~\ref{fig:calvin_ex2}~\&~\ref{fig:calvin_ex3}. 

\newpage

\begin{figure}[ht!]
    \centering
        \includegraphics[width=\columnwidth, trim={0.25in 1.75in 1.75in 4.75in},clip]{figures_app/vb_ex1.pdf}
        \caption{An example of an L4 generalization level task (novel task generalization of ``same texture'') in VIMA-Bench.}
        \label{fig:vb_ex2}
\end{figure}






\begin{figure}[ht!]
    \centering
        \includegraphics[width=\columnwidth, trim={0.25in 2.125in 1.75in 4.75in},clip]{figures_app/calvin_ex1.pdf}
        \caption{An example of a task in CALVIN.}
        \label{fig:calvin_ex2}
\end{figure}

%% file: tables/notation.tex
\begin{table}[h]
    \centering
    \small
    \begin{tabular}{p{0.14\columnwidth} p{0.75\columnwidth}}
        \toprule
        \textbf{Symbol} & \textbf{Description} \\

        \midrule
        \multicolumn{2}{l}{\textit{Basic symbols}} \\
        $\tau$ & EIF trajectory \\
        $p$ & Language instruction/prompt \\
        $s$ & State (visual observation) \\
        $a$ & Action \\
        $q$ & Action queries (shared across timesteps) \\

        \midrule
        \multicolumn{2}{l}{\textit{Segment-level symbols}} \\
        $s_{t}$ & State segment at timestep $t$ \\
        $a_{t}$ & Action segment at timestep $t$ \\
        $s_{\leq t}$ & All states up to and including timestep $t$ \\
        $a_{< t}$ & All actions before timestep $t$ \\
        $t$ & Timestep index ($1 \leq t \leq T$) \\
        $T$ & Total number of timesteps in a trajectory \\

        \midrule
        \multicolumn{2}{l}{\textit{Token-level symbols}} \\
        $p_{i}$ & The $i$-th prompt token \\
        $s_{i,t}$ & The $i$-th state token at timestep $t$ \\
        $a_{i,t}$ & The $i$-th action dimension at timestep $t$ \\
        $q_{i}$ / $q_{i,t}$ & The $i$-th action query (shared across timesteps) \\

        \midrule
        \multicolumn{2}{l}{\textit{Token slices}} \\
        $p_{1:L}$ & Prompt tokens $1$ through $L$ \\
        $s_{1:M,t}$ & State tokens $1$ through $M$ at timestep $t$ \\
        $a_{1:N,t}$ & Action dimensions $1$ through $N$ at timestep $t$ \\
        $q_{1:N}$ & Action queries $1$ through $N$ (shared across timesteps) \\
        $L$ & Number of language instruction tokens \\
        $M$ & Number of state tokens per segment \\
        $N$ & Number of action dimensions/queries per segment \\

        \midrule
        \multicolumn{2}{l}{\textit{Contrastive dynamics learning}} \\
        $\tau^+$ & Positive trajectory sample \\
        $\tau^-$ & Negative trajectory sample \\
        $\tau', \tau''$ & Alternative trajectories different from $\tau$ \\

        \midrule
        \multicolumn{2}{l}{\textit{Model}} \\
        $\pi$ & Policy \\
        $\pi_\theta$ & Policy parameterized by $\theta$ \\
        $\theta$ & Model parameters \\
        $f(\cdot)$ & Trajectory encoding model \\

        \bottomrule
    \end{tabular}
\end{table}

%% file: tables/franka_kitchen.tex
\begin{table*}
\centering
    \caption{Performance comparison (\%) in Franka Kitchen.}
    \begin{tabular}{l | c c | c c c c c | c c c c c}
    \toprule
    \multicolumn{1}{c|}{} & \multicolumn{2}{c|}{\textbf{Configuration}} & \multicolumn{5}{c|}{Continuous Action} & \multicolumn{5}{c}{Discrete Action} \\

        \textbf{Model} & \textbf{Attn Type} & \textbf{Params} & $p\text{1}$ & $p\text{2}$ & $p\text{3}$ & $p\text{4}$ & \textbf{Mean} & $p\text{1}$ & $p\text{2}$ & $p\text{3}$ & $p\text{4}$ & \textbf{Mean} \\

        \midrule
        Diffusion & Causal & 43M  & \textbf{100}  & 94 & 72 & 34 & 76 &
                                        54 & 18 & 10 & 2 & 21 \\

        VIMA &      Cross  & 113M  & 92 & 90 & 80 & 66 & 82 &
                                        90 & 72& 48 & 16 & 57 \\

        Gato &      Causal &  43M  & \textbf{100}  & 98 & 84 & 62& 86 &
                                        \textbf{100}  & 88& \textbf{68} & 38 & 74 \\

        \hline
        Astra (ours) &     Traj. &  43M  & \textbf{100}  & \textbf{100} & \textbf{94} & \textbf{70} & \textbf{91} &
                                        \textbf{100}  & \textbf{94} & \textbf{68} & \textbf{52} & \textbf{79} \\

    \bottomrule
    \end{tabular}
    \label{tab:franka_kitchen}
\end{table*}

%% file: tables/push_T.tex
\begin{table}
\centering
    \caption{Performance comparison in Push-T (Discrete Action).}
    \begin{tabular}{l | c c | c}
    \toprule

        \textbf{Model} & \textbf{Attn} & \textbf{Params} & \textbf{Score} \\

        \midrule
        Diffusion & Causal & 43M &  83.89 \\

        VIMA &      Cross & 26M  &  91.09 \\

        Gato &      Causal & 19M  &  90.43 \\

        \hline
        Astra (ours) & Traj. & 19M  & \textbf{94.11} \\

    \bottomrule
    \end{tabular}
    \label{tab:push_T}
\end{table}

%% file: main.bbl
\begin{thebibliography}{68}
\providecommand{\natexlab}[1]{#1}

\bibitem[{Alayrac et~al.(2022)Alayrac, Donahue, Luc, Miech, Barr, Hasson, Lenc,
  Mensch, Millican, Reynolds, Ring, Rutherford, Cabi, Han, Gong, Samangooei,
  Monteiro, Menick, Borgeaud, Brock, Nematzadeh, Sharifzadeh, Binkowski,
  Barreira, Vinyals, Zisserman, and
  Simonyan}]{DBLP:conf/nips/AlayracDLMBHLMM22}
Jean{-}Baptiste Alayrac, Jeff Donahue, Pauline Luc, Antoine Miech, Iain Barr,
  Yana Hasson, Karel Lenc, Arthur Mensch, Katherine Millican, Malcolm Reynolds,
  Roman Ring, Eliza Rutherford, Serkan Cabi, Tengda Han, Zhitao Gong, Sina
  Samangooei, Marianne Monteiro, Jacob~L. Menick, Sebastian Borgeaud, and 8
  others. 2022.
\newblock Flamingo: a visual language model for few-shot learning.
\newblock In \emph{NeurIPS}.

\bibitem[{Austin et~al.(2021)Austin, Johnson, Ho, Tarlow, and van~den
  Berg}]{DBLP:conf/nips/AustinJHTB21}
Jacob Austin, Daniel~D. Johnson, Jonathan Ho, Daniel Tarlow, and Rianne van~den
  Berg. 2021.
\newblock Structured denoising diffusion models in discrete state-spaces.
\newblock In \emph{NeurIPS}, pages 17981--17993.

\bibitem[{Awadalla et~al.(2023)Awadalla, Gao, Gardner, Hessel, Hanafy, Zhu,
  Marathe, Bitton, Gadre, Sagawa, Jitsev, Kornblith, Koh, Ilharco, Wortsman,
  and Schmidt}]{DBLP:journals/corr/abs-2308-01390}
Anas Awadalla, Irena Gao, Josh Gardner, Jack Hessel, Yusuf Hanafy, Wanrong Zhu,
  Kalyani Marathe, Yonatan Bitton, Samir~Yitzhak Gadre, Shiori Sagawa, Jenia
  Jitsev, Simon Kornblith, Pang~Wei Koh, Gabriel Ilharco, Mitchell Wortsman,
  and Ludwig Schmidt. 2023.
\newblock Openflamingo: An open-source framework for training large
  autoregressive vision-language models.
\newblock \emph{CoRR}, abs/2308.01390.

\bibitem[{Bai et~al.(2025)Bai, Chen, Liu, Wang, Ge, Song, Dang, Wang, Wang,
  Tang, Zhong, Zhu, Yang, Li, Wan, Wang, Ding, Fu, Xu, Ye, Zhang, Xie, Cheng,
  Zhang, Yang, Xu, and Lin}]{DBLP:journals/corr/abs-2502-13923}
Shuai Bai, Keqin Chen, Xuejing Liu, Jialin Wang, Wenbin Ge, Sibo Song, Kai
  Dang, Peng Wang, Shijie Wang, Jun Tang, Humen Zhong, Yuanzhi Zhu,
  Ming{-}Hsuan Yang, Zhaohai Li, Jianqiang Wan, Pengfei Wang, Wei Ding, Zheren
  Fu, Yiheng Xu, and 8 others. 2025.
\newblock Qwen2.5-vl technical report.
\newblock \emph{CoRR}, abs/2502.13923.

\bibitem[{Beyer et~al.(2024)Beyer, Steiner, Pinto, Kolesnikov, Wang, Salz,
  Neumann, Alabdulmohsin, Tschannen, Bugliarello, Unterthiner, Keysers,
  Koppula, Liu, Grycner, Gritsenko, Houlsby, Kumar, Rong, Eisenschlos, Kabra,
  Bauer, Bosnjak, Chen, Minderer, Voigtlaender, Bica, Balazevic, Puigcerver,
  Papalampidi, H{\'{e}}naff, Xiong, Soricut, Harmsen, and
  Zhai}]{DBLP:journals/corr/abs-2407-07726}
Lucas Beyer, Andreas Steiner, Andr{\'{e}}~Susano Pinto, Alexander Kolesnikov,
  Xiao Wang, Daniel Salz, Maxim Neumann, Ibrahim Alabdulmohsin, Michael
  Tschannen, Emanuele Bugliarello, Thomas Unterthiner, Daniel Keysers, Skanda
  Koppula, Fangyu Liu, Adam Grycner, Alexey~A. Gritsenko, Neil Houlsby, Manoj
  Kumar, Keran Rong, and 16 others. 2024.
\newblock Paligemma: {A} versatile 3b {VLM} for transfer.
\newblock \emph{CoRR}, abs/2407.07726.

\bibitem[{Black et~al.(2024)Black, Brown, Driess, Esmail, Equi, Finn, Fusai,
  Groom, Hausman, Ichter, Jakubczak, Jones, Ke, Levine, Li{-}Bell, Mothukuri,
  Nair, Pertsch, Shi, Tanner, Vuong, Walling, Wang, and
  Zhilinsky}]{DBLP:journals/corr/abs-2410-24164}
Kevin Black, Noah Brown, Danny Driess, Adnan Esmail, Michael Equi, Chelsea
  Finn, Niccolo Fusai, Lachy Groom, Karol Hausman, Brian Ichter, Szymon
  Jakubczak, Tim Jones, Liyiming Ke, Sergey Levine, Adrian Li{-}Bell, Mohith
  Mothukuri, Suraj Nair, Karl Pertsch, Lucy~Xiaoyang Shi, and 5 others. 2024.
\newblock {\(\pi\)}\({}_{\mbox{0}}\): {A} vision-language-action flow model for
  general robot control.
\newblock \emph{CoRR}, abs/2410.24164.

\bibitem[{Brohan et~al.(2023{\natexlab{a}})Brohan, Brown, Carbajal, Chebotar,
  Chen, Choromanski, Ding, Driess, Dubey, Finn, Florence, Fu, Arenas,
  Gopalakrishnan, Han, Hausman, Herzog, Hsu, Ichter, Irpan, Joshi, Julian,
  Kalashnikov, Kuang, Leal, Lee, Lee, Levine, Lu, Michalewski, Mordatch,
  Pertsch, Rao, Reymann, Ryoo, Salazar, Sanketi, Sermanet, Singh, Singh,
  Soricut, Tran, Vanhoucke, Vuong, Wahid, Welker, Wohlhart, Wu, Xia, Xiao, Xu,
  Xu, Yu, and Zitkovich}]{DBLP:journals/corr/abs-2307-15818}
Anthony Brohan, Noah Brown, Justice Carbajal, Yevgen Chebotar, Xi~Chen,
  Krzysztof Choromanski, Tianli Ding, Danny Driess, Avinava Dubey, Chelsea
  Finn, Pete Florence, Chuyuan Fu, Montse~Gonzalez Arenas, Keerthana
  Gopalakrishnan, Kehang Han, Karol Hausman, Alexander Herzog, Jasmine Hsu,
  Brian Ichter, and 35 others. 2023{\natexlab{a}}.
\newblock {RT-2:} vision-language-action models transfer web knowledge to
  robotic control.
\newblock \emph{CoRR}, abs/2307.15818.

\bibitem[{Brohan et~al.(2023{\natexlab{b}})Brohan, Brown, Carbajal, Chebotar,
  Dabis, Finn, Gopalakrishnan, Hausman, Herzog, Hsu, Ibarz, Ichter, Irpan,
  Jackson, Jesmonth, Joshi, Julian, Kalashnikov, Kuang, Leal, Lee, Levine, Lu,
  Malla, Manjunath, Mordatch, Nachum, Parada, Peralta, Perez, Pertsch,
  Quiambao, Rao, Ryoo, Salazar, Sanketi, Sayed, Singh, Sontakke, Stone, Tan,
  Tran, Vanhoucke, Vega, Vuong, Xia, Xiao, Xu, Xu, Yu, and
  Zitkovich}]{DBLP:conf/rss/BrohanBCCDFGHHH23}
Anthony Brohan, Noah Brown, Justice Carbajal, Yevgen Chebotar, Joseph Dabis,
  Chelsea Finn, Keerthana Gopalakrishnan, Karol Hausman, Alexander Herzog,
  Jasmine Hsu, Julian Ibarz, Brian Ichter, Alex Irpan, Tomas Jackson, Sally
  Jesmonth, Nikhil~J. Joshi, Ryan Julian, Dmitry Kalashnikov, Yuheng Kuang, and
  32 others. 2023{\natexlab{b}}.
\newblock {RT-1:} robotics transformer for real-world control at scale.
\newblock In \emph{Robotics: Science and Systems}.

\bibitem[{Carion et~al.(2020)Carion, Massa, Synnaeve, Usunier, Kirillov, and
  Zagoruyko}]{DBLP:conf/eccv/CarionMSUKZ20}
Nicolas Carion, Francisco Massa, Gabriel Synnaeve, Nicolas Usunier, Alexander
  Kirillov, and Sergey Zagoruyko. 2020.
\newblock End-to-end object detection with transformers.
\newblock In \emph{{ECCV} {(1)}}, volume 12346 of \emph{Lecture Notes in
  Computer Science}, pages 213--229. Springer.

\bibitem[{Cheang et~al.(2024)Cheang, Chen, Jing, Kong, Li, Li, Liu, Wu, Xu,
  Yang, Zhang, and Zhu}]{DBLP:journals/corr/abs-2410-06158}
Chilam Cheang, Guangzeng Chen, Ya~Jing, Tao Kong, Hang Li, Yifeng Li, Yuxiao
  Liu, Hongtao Wu, Jiafeng Xu, Yichu Yang, Hanbo Zhang, and Minzhao Zhu. 2024.
\newblock {GR-2:} {A} generative video-language-action model with web-scale
  knowledge for robot manipulation.
\newblock \emph{CoRR}, abs/2410.06158.

\bibitem[{Chebotar et~al.(2023)Chebotar, Vuong, Irpan, Hausman, Xia, Lu, Kumar,
  Yu, Herzog, Pertsch, Gopalakrishnan, Ibarz, Nachum, Sontakke, Salazar, Tran,
  Peralta, Tan, Manjunath, Singh, Zitkovich, Jackson, Rao, Finn, and
  Levine}]{DBLP:journals/corr/abs-2309-10150}
Yevgen Chebotar, Quan Vuong, Alex Irpan, Karol Hausman, Fei Xia, Yao Lu, Aviral
  Kumar, Tianhe Yu, Alexander Herzog, Karl Pertsch, Keerthana Gopalakrishnan,
  Julian Ibarz, Ofir Nachum, Sumedh Sontakke, Grecia Salazar, Huong~T. Tran,
  Jodilyn Peralta, Clayton Tan, Deeksha Manjunath, and 6 others. 2023.
\newblock Q-transformer: Scalable offline reinforcement learning via
  autoregressive q-functions.
\newblock \emph{CoRR}, abs/2309.10150.

\bibitem[{Chen et~al.(2021)Chen, Lu, Rajeswaran, Lee, Grover, Laskin, Abbeel,
  Srinivas, and Mordatch}]{DBLP:conf/nips/ChenLRLGLASM21}
Lili Chen, Kevin Lu, Aravind Rajeswaran, Kimin Lee, Aditya Grover, Michael
  Laskin, Pieter Abbeel, Aravind Srinivas, and Igor Mordatch. 2021.
\newblock Decision transformer: Reinforcement learning via sequence modeling.
\newblock In \emph{NeurIPS}, pages 15084--15097.

\bibitem[{Chen et~al.(2024)Chen, Luo, Zhang, and
  Tian}]{DBLP:journals/mta/ChenLZT24}
Wei Chen, Jinjin Luo, Fan Zhang, and Zijian Tian. 2024.
\newblock A review of object detection: Datasets, performance evaluation,
  architecture, applications and current trends.
\newblock \emph{Multim. Tools Appl.}, 83(24):65603--65661.

\bibitem[{Devlin et~al.(2019)Devlin, Chang, Lee, and
  Toutanova}]{DBLP:conf/naacl/DevlinCLT19}
Jacob Devlin, Ming{-}Wei Chang, Kenton Lee, and Kristina Toutanova. 2019.
\newblock {BERT:} pre-training of deep bidirectional transformers for language
  understanding.
\newblock In \emph{{NAACL-HLT} {(1)}}, pages 4171--4186. Association for
  Computational Linguistics.

\bibitem[{Dong et~al.(2019)Dong, Yang, Wang, Wei, Liu, Wang, Gao, Zhou, and
  Hon}]{DBLP:conf/nips/00040WWLWGZH19}
Li~Dong, Nan Yang, Wenhui Wang, Furu Wei, Xiaodong Liu, Yu~Wang, Jianfeng Gao,
  Ming Zhou, and Hsiao{-}Wuen Hon. 2019.
\newblock Unified language model pre-training for natural language
  understanding and generation.
\newblock In \emph{NeurIPS}, pages 13042--13054.

\bibitem[{Dosovitskiy et~al.(2021)Dosovitskiy, Beyer, Kolesnikov, Weissenborn,
  Zhai, Unterthiner, Dehghani, Minderer, Heigold, Gelly, Uszkoreit, and
  Houlsby}]{DBLP:conf/iclr/DosovitskiyB0WZ21}
Alexey Dosovitskiy, Lucas Beyer, Alexander Kolesnikov, Dirk Weissenborn,
  Xiaohua Zhai, Thomas Unterthiner, Mostafa Dehghani, Matthias Minderer, Georg
  Heigold, Sylvain Gelly, Jakob Uszkoreit, and Neil Houlsby. 2021.
\newblock An image is worth 16x16 words: Transformers for image recognition at
  scale.
\newblock In \emph{{ICLR}}. OpenReview.net.

\bibitem[{Driess et~al.(2023)Driess, Xia, Sajjadi, Lynch, Chowdhery, Ichter,
  Wahid, Tompson, Vuong, Yu, Huang, Chebotar, Sermanet, Duckworth, Levine,
  Vanhoucke, Hausman, Toussaint, Greff, Zeng, Mordatch, and
  Florence}]{DBLP:conf/icml/DriessXSLCIWTVY23}
Danny Driess, Fei Xia, Mehdi S.~M. Sajjadi, Corey Lynch, Aakanksha Chowdhery,
  Brian Ichter, Ayzaan Wahid, Jonathan Tompson, Quan Vuong, Tianhe Yu, Wenlong
  Huang, Yevgen Chebotar, Pierre Sermanet, Daniel Duckworth, Sergey Levine,
  Vincent Vanhoucke, Karol Hausman, Marc Toussaint, Klaus Greff, and 3 others.
  2023.
\newblock Palm-e: An embodied multimodal language model.
\newblock In \emph{{ICML}}, volume 202 of \emph{Proceedings of Machine Learning
  Research}, pages 8469--8488. {PMLR}.

\bibitem[{Du et~al.(2023)Du, Yang, Dai, Dai, Nachum, Tenenbaum, Schuurmans, and
  Abbeel}]{DBLP:conf/nips/DuY0DN0SA23}
Yilun Du, Sherry Yang, Bo~Dai, Hanjun Dai, Ofir Nachum, Josh Tenenbaum, Dale
  Schuurmans, and Pieter Abbeel. 2023.
\newblock Learning universal policies via text-guided video generation.
\newblock In \emph{NeurIPS}.

\bibitem[{Florence et~al.(2021)Florence, Lynch, Zeng, Ramirez, Wahid, Downs,
  Wong, Lee, Mordatch, and Tompson}]{DBLP:conf/corl/FlorenceLZRWDWL21}
Pete Florence, Corey Lynch, Andy Zeng, Oscar~A. Ramirez, Ayzaan Wahid, Laura
  Downs, Adrian Wong, Johnny Lee, Igor Mordatch, and Jonathan Tompson. 2021.
\newblock Implicit behavioral cloning.
\newblock In \emph{CoRL}, volume 164 of \emph{Proceedings of Machine Learning
  Research}, pages 158--168. {PMLR}.

\bibitem[{Ghosh et~al.(2024)Ghosh, Acharya, Saha, Jain, and
  Chadha}]{DBLP:journals/corr/abs-2404-07214}
Akash Ghosh, Arkadeep Acharya, Sriparna Saha, Vinija Jain, and Aman Chadha.
  2024.
\newblock \href {https://doi.org/10.48550/ARXIV.2404.07214} {Exploring the
  frontier of vision-language models: {A} survey of current methodologies and
  future directions}.
\newblock \emph{CoRR}, abs/2404.07214.

\bibitem[{Gu et~al.(2023)Gu, Xiang, Li, Ling, Liu, Mu, Tang, Tao, Wei, Yao,
  Yuan, Xie, Huang, Chen, and Su}]{gu2023maniskill2}
Jiayuan Gu, Fanbo Xiang, Xuanlin Li, Zhan Ling, Xiqiang Liu, Tongzhou Mu, Yihe
  Tang, Stone Tao, Xinyue Wei, Yunchao Yao, Xiaodi Yuan, Pengwei Xie, Zhiao
  Huang, Rui Chen, and Hao Su. 2023.
\newblock Maniskill2: A unified benchmark for generalizable manipulation
  skills.
\newblock In \emph{International Conference on Learning Representations}.

\bibitem[{Gupta et~al.(2019)Gupta, Kumar, Lynch, Levine, and
  Hausman}]{DBLP:conf/corl/0004KLLH19}
Abhishek Gupta, Vikash Kumar, Corey Lynch, Sergey Levine, and Karol Hausman.
  2019.
\newblock Relay policy learning: Solving long-horizon tasks via imitation and
  reinforcement learning.
\newblock In \emph{3rd Annual Conference on Robot Learning, CoRL 2019, Osaka,
  Japan, October 30 - November 1, 2019, Proceedings}, volume 100 of
  \emph{Proceedings of Machine Learning Research}, pages 1025--1037. {PMLR}.

\bibitem[{Hafner et~al.(2020)Hafner, Lillicrap, Ba, and
  Norouzi}]{DBLP:conf/iclr/HafnerLB020}
Danijar Hafner, Timothy~P. Lillicrap, Jimmy Ba, and Mohammad Norouzi. 2020.
\newblock Dream to control: Learning behaviors by latent imagination.
\newblock In \emph{{ICLR}}. OpenReview.net.

\bibitem[{He et~al.(2022)He, Chen, Xie, Li, Doll{\'{a}}r, and
  Girshick}]{DBLP:conf/cvpr/HeCXLDG22}
Kaiming He, Xinlei Chen, Saining Xie, Yanghao Li, Piotr Doll{\'{a}}r, and
  Ross~B. Girshick. 2022.
\newblock Masked autoencoders are scalable vision learners.
\newblock In \emph{{CVPR}}, pages 15979--15988. {IEEE}.

\bibitem[{He et~al.(2016)He, Zhang, Ren, and Sun}]{DBLP:conf/cvpr/HeZRS16}
Kaiming He, Xiangyu Zhang, Shaoqing Ren, and Jian Sun. 2016.
\newblock Deep residual learning for image recognition.
\newblock In \emph{{CVPR}}, pages 770--778. {IEEE} Computer Society.

\bibitem[{Ho et~al.(2020)Ho, Jain, and Abbeel}]{DBLP:conf/nips/HoJA20}
Jonathan Ho, Ajay Jain, and Pieter Abbeel. 2020.
\newblock Denoising diffusion probabilistic models.
\newblock In \emph{NeurIPS}.

\bibitem[{Ichter et~al.(2022)Ichter, Brohan, Chebotar, Finn, Hausman, Herzog,
  Ho, Ibarz, Irpan, Jang, Julian, Kalashnikov, Levine, Lu, Parada, Rao,
  Sermanet, Toshev, Vanhoucke, Xia, Xiao, Xu, Yan, Brown, Ahn, Cortes, Sievers,
  Tan, Xu, Reyes, Rettinghouse, Quiambao, Pastor, Luu, Lee, Kuang, Jesmonth,
  Joshi, Jeffrey, Ruano, Hsu, Gopalakrishnan, David, Zeng, and
  Fu}]{DBLP:conf/corl/IchterBCFHHHIIJ22}
Brian Ichter, Anthony Brohan, Yevgen Chebotar, Chelsea Finn, Karol Hausman,
  Alexander Herzog, Daniel Ho, Julian Ibarz, Alex Irpan, Eric Jang, Ryan
  Julian, Dmitry Kalashnikov, Sergey Levine, Yao Lu, Carolina Parada, Kanishka
  Rao, Pierre Sermanet, Alexander Toshev, Vincent Vanhoucke, and 26 others.
  2022.
\newblock Do as {I} can, not as {I} say: Grounding language in robotic
  affordances.
\newblock In \emph{CoRL}, volume 205 of \emph{Proceedings of Machine Learning
  Research}, pages 287--318. {PMLR}.

\bibitem[{Janner et~al.(2021)Janner, Li, and
  Levine}]{DBLP:conf/nips/JannerLL21}
Michael Janner, Qiyang Li, and Sergey Levine. 2021.
\newblock Offline reinforcement learning as one big sequence modeling problem.
\newblock In \emph{NeurIPS}, pages 1273--1286.

\bibitem[{Jiang et~al.(2022)Jiang, Gupta, Zhang, Wang, Dou, Chen, Fei{-}Fei,
  Anandkumar, Zhu, and Fan}]{DBLP:journals/corr/abs-2210-03094}
Yunfan Jiang, Agrim Gupta, Zichen Zhang, Guanzhi Wang, Yongqiang Dou, Yanjun
  Chen, Li~Fei{-}Fei, Anima Anandkumar, Yuke Zhu, and Linxi Fan. 2022.
\newblock {VIMA:} general robot manipulation with multimodal prompts.
\newblock \emph{CoRR}, abs/2210.03094.

\bibitem[{Karamcheti et~al.(2023)Karamcheti, Nair, Chen, Kollar, Finn, Sadigh,
  and Liang}]{DBLP:conf/rss/KaramchetiNCKFS23}
Siddharth Karamcheti, Suraj Nair, Annie~S. Chen, Thomas Kollar, Chelsea Finn,
  Dorsa Sadigh, and Percy Liang. 2023.
\newblock Language-driven representation learning for robotics.
\newblock In \emph{Robotics: Science and Systems}.

\bibitem[{Kim et~al.(2024)Kim, Pertsch, Karamcheti, Xiao, Balakrishna, Nair,
  Rafailov, Foster, Lam, Sanketi, Vuong, Kollar, Burchfiel, Tedrake, Sadigh,
  Levine, Liang, and Finn}]{DBLP:journals/corr/abs-2406-09246}
Moo~Jin Kim, Karl Pertsch, Siddharth Karamcheti, Ted Xiao, Ashwin Balakrishna,
  Suraj Nair, Rafael Rafailov, Ethan~Paul Foster, Grace Lam, Pannag Sanketi,
  Quan Vuong, Thomas Kollar, Benjamin Burchfiel, Russ Tedrake, Dorsa Sadigh,
  Sergey Levine, Percy Liang, and Chelsea Finn. 2024.
\newblock Openvla: An open-source vision-language-action model.
\newblock \emph{CoRR}, abs/2406.09246.

\bibitem[{Li et~al.(2024)Li, Gao, Johnston, Gao, He, Shi, Shakiah, Ghanadan,
  and Wang}]{DBLP:conf/icml/LiGJ0HSSGW24}
Jiachen Li, Qiaozi Gao, Michael Johnston, Xiaofeng Gao, Xuehai He, Hangjie Shi,
  Suhaila Shakiah, Reza Ghanadan, and William~Yang Wang. 2024.
\newblock Mastering robot manipulation with multimodal prompts through
  pretraining and multi-task fine-tuning.
\newblock In \emph{{ICML}}. OpenReview.net.

\bibitem[{Li et~al.(2023{\natexlab{a}})Li, Li, Savarese, and
  Hoi}]{DBLP:conf/icml/0008LSH23}
Junnan Li, Dongxu Li, Silvio Savarese, and Steven C.~H. Hoi.
  2023{\natexlab{a}}.
\newblock {BLIP-2:} bootstrapping language-image pre-training with frozen image
  encoders and large language models.
\newblock In \emph{{ICML}}, volume 202 of \emph{Proceedings of Machine Learning
  Research}, pages 19730--19742. {PMLR}.

\bibitem[{Li et~al.(2023{\natexlab{b}})Li, Liu, Zhang, Yu, Xu, Wu, Cheang,
  Jing, Zhang, Liu, Li, and Kong}]{DBLP:journals/corr/abs-2311-01378}
Xinghang Li, Minghuan Liu, Hanbo Zhang, Cunjun Yu, Jie Xu, Hongtao Wu, Chilam
  Cheang, Ya~Jing, Weinan Zhang, Huaping Liu, Hang Li, and Tao Kong.
  2023{\natexlab{b}}.
\newblock Vision-language foundation models as effective robot imitators.
\newblock \emph{CoRR}, abs/2311.01378.

\bibitem[{Lin et~al.(2021)Lin, Men, Yang, Zhou, Ding, Zhang, Wang, Wang, Jiang,
  Jia, Zhang, Zhang, Zou, Li, Deng, Liu, Xue, Zhou, Ma, Yu, Li, Lin, Zhou,
  Tang, and Yang}]{DBLP:journals/corr/abs-2103-00823}
Junyang Lin, Rui Men, An~Yang, Chang Zhou, Ming Ding, Yichang Zhang, Peng Wang,
  Ang Wang, Le~Jiang, Xianyan Jia, Jie Zhang, Jianwei Zhang, Xu~Zou, Zhikang
  Li, Xiaodong Deng, Jie Liu, Jinbao Xue, Huiling Zhou, Jianxin Ma, and 6
  others. 2021.
\newblock {M6:} {A} chinese multimodal pretrainer.
\newblock \emph{CoRR}, abs/2103.00823.

\bibitem[{Liu et~al.(2022)Liu, Liu, Grover, and
  Abbeel}]{DBLP:conf/nips/Liu0GA22}
Fangchen Liu, Hao Liu, Aditya Grover, and Pieter Abbeel. 2022.
\newblock Masked autoencoding for scalable and generalizable decision making.
\newblock In \emph{NeurIPS}.

\bibitem[{Loshchilov and Hutter(2019)}]{DBLP:conf/iclr/LoshchilovH19}
Ilya Loshchilov and Frank Hutter. 2019.
\newblock Decoupled weight decay regularization.
\newblock In \emph{{ICLR} (Poster)}. OpenReview.net.

\bibitem[{Lu et~al.(2025)Lu, Wang, Liu, Lu, and Tang}]{DBLP:conf/iclr/LuWLLT25}
Guanxing Lu, Ziwei Wang, Changliu Liu, Jiwen Lu, and Yansong Tang. 2025.
\newblock Thinkbot: Embodied instruction following with thought chain
  reasoning.
\newblock In \emph{{ICLR}}. OpenReview.net.

\bibitem[{Lu et~al.(2019)Lu, Batra, Parikh, and Lee}]{DBLP:conf/nips/LuBPL19}
Jiasen Lu, Dhruv Batra, Devi Parikh, and Stefan Lee. 2019.
\newblock Vilbert: Pretraining task-agnostic visiolinguistic representations
  for vision-and-language tasks.
\newblock In \emph{NeurIPS}, pages 13--23.

\bibitem[{Ma et~al.(2023{\natexlab{a}})Ma, Sodhani, Jayaraman, Bastani, Kumar,
  and Zhang}]{DBLP:conf/iclr/MaSJBK023}
Yecheng~Jason Ma, Shagun Sodhani, Dinesh Jayaraman, Osbert Bastani, Vikash
  Kumar, and Amy Zhang. 2023{\natexlab{a}}.
\newblock {VIP:} towards universal visual reward and representation via
  value-implicit pre-training.
\newblock In \emph{{ICLR}}. OpenReview.net.

\bibitem[{Ma et~al.(2024{\natexlab{a}})Ma, Chi, Li, Song, Zhuang, and
  King}]{DBLP:conf/naacl/MaCLSZK24}
Yueen Ma, Dafeng Chi, Jingjing Li, Kai Song, Yuzheng Zhuang, and Irwin King.
  2024{\natexlab{a}}.
\newblock {VOLTA:} improving generative diversity by variational mutual
  information maximizing autoencoder.
\newblock In \emph{{NAACL-HLT} (Findings)}, pages 364--378. Association for
  Computational Linguistics.

\bibitem[{Ma et~al.(2023{\natexlab{b}})Ma, Song, Hu, Li, Zhang, and
  King}]{DBLP:conf/aaai/MaSHLZK23}
Yueen Ma, Zixing Song, Xuming Hu, Jingjing Li, Yifei Zhang, and Irwin King.
  2023{\natexlab{b}}.
\newblock Graph component contrastive learning for concept relatedness
  estimation.
\newblock In \emph{{AAAI}}, pages 13362--13370. {AAAI} Press.

\bibitem[{Ma et~al.(2024{\natexlab{b}})Ma, Song, Zhuang, Hao, and
  King}]{DBLP:journals/corr/abs-2405-14093}
Yueen Ma, Zixing Song, Yuzheng Zhuang, Jianye Hao, and Irwin King.
  2024{\natexlab{b}}.
\newblock A survey on vision-language-action models for embodied {AI}.
\newblock \emph{CoRR}, abs/2405.14093.

\bibitem[{Ma et~al.(2021)Ma, Yan, Long, Rangaprakash, and
  Deshpande}]{DBLP:conf/ijcnn/MaYLRD21}
Yueen Ma, Da~Yan, Cheng Long, D.~Rangaprakash, and Gopikrishna Deshpande. 2021.
\newblock Predicting autism spectrum disorder from brain imaging data by graph
  convolutional network.
\newblock In \emph{{IJCNN}}, pages 1--8. {IEEE}.

\bibitem[{Mees et~al.(2022)Mees, Hermann, Rosete{-}Beas, and
  Burgard}]{DBLP:journals/ral/MeesHRB22}
Oier Mees, Luk{\'{a}}s Hermann, Erick Rosete{-}Beas, and Wolfram Burgard. 2022.
\newblock {CALVIN:} {A} benchmark for language-conditioned policy learning for
  long-horizon robot manipulation tasks.
\newblock \emph{{IEEE} Robotics Autom. Lett.}, 7(3):7327--7334.

\bibitem[{Micheli et~al.(2023)Micheli, Alonso, and
  Fleuret}]{DBLP:conf/iclr/MicheliAF23}
Vincent Micheli, Eloi Alonso, and Fran{\c{c}}ois Fleuret. 2023.
\newblock Transformers are sample-efficient world models.
\newblock In \emph{{ICLR}}. OpenReview.net.

\bibitem[{Nair et~al.(2022)Nair, Rajeswaran, Kumar, Finn, and
  Gupta}]{DBLP:conf/corl/NairRKF022}
Suraj Nair, Aravind Rajeswaran, Vikash Kumar, Chelsea Finn, and Abhinav Gupta.
  2022.
\newblock {R3M:} {A} universal visual representation for robot manipulation.
\newblock In \emph{CoRL}, volume 205 of \emph{Proceedings of Machine Learning
  Research}, pages 892--909. {PMLR}.

\bibitem[{O'Neill et~al.(2024)O'Neill, Rehman, Maddukuri, Gupta, Padalkar, Lee,
  Pooley, Gupta, Mandlekar, Jain, Tung, Bewley, Herzog, Irpan, Khazatsky, Rai,
  Gupta, Wang, Singh, Garg, Kembhavi, Xie, Brohan, Raffin, Sharma, Yavary,
  Jain, Balakrishna, Wahid, Burgess{-}Limerick, Kim, Sch{\"{o}}lkopf, Wulfe,
  Ichter, Lu, Xu, Le, Finn, Wang, Xu, Chi, Huang, Chan, Agia, Pan, Fu, Devin,
  Xu, Morton, Driess, Chen, Pathak, Shah, B{\"{u}}chler, Jayaraman,
  Kalashnikov, Sadigh, Johns, Foster, Liu, Ceola, Xia, Zhao, Stulp, Zhou,
  Sukhatme, Salhotra, Yan, Feng, Schiavi, Berseth, Kahn, Wang, Su, Fang, Shi,
  Bao, Amor, Christensen, Furuta, Walke, Fang, Ha, Mordatch, Radosavovic, Leal,
  Liang, Abou{-}Chakra, Kim, Drake, Peters, Schneider, Hsu, Bohg, Bingham, Wu,
  Gao, Hu, Wu, Wu, Sun, Luo, Gu, Tan, Oh, Wu, Lu, Yang, Malik, Silv{\'{e}}rio,
  Hejna, Booher, Tompson, Yang, Salvador, Lim, Han, Wang, Rao, Pertsch,
  Hausman, Go, Gopalakrishnan, Goldberg, Byrne, Oslund, Kawaharazuka, Black,
  Lin, Zhang, Ehsani, Lekkala, Ellis, Rana, Srinivasan, Fang, Singh, Zeng,
  Hatch, Hsu, Itti, Chen, Pinto, Fei{-}Fei, Tan, Fan, Ott, Lee, Weihs, Chen,
  Lepert, Memmel, Tomizuka, Itkina, Castro, Spero, Du, Ahn, Yip, Zhang, Ding,
  Heo, Srirama, Sharma, Kim, Kanazawa, Hansen, Heess, Joshi, S{\"{u}}nderhauf,
  Liu, Palo, Shafiullah, Mees, Kroemer, Bastani, Sanketi, Miller, Yin,
  Wohlhart, Xu, Fagan, Mitrano, Sermanet, Abbeel, Sundaresan, Chen, Vuong,
  Rafailov, Tian, Doshi, Mart{\'{\i}}n{-}Mart{\'{\i}}n, Baijal, Scalise,
  Hendrix, Lin, Qian, Zhang, Mendonca, Shah, Hoque, Julian, Bustamante,
  Kirmani, Levine, Lin, Moore, Bahl, Dass, Sonawani, Song, Xu, Haldar,
  Karamcheti, Adebola, Guist, Nasiriany, Schaal, Welker, Tian, Ramamoorthy,
  Dasari, Belkhale, Park, Nair, Mirchandani, Osa, Gupta, Harada, Matsushima,
  Xiao, Kollar, Yu, Ding, Davchev, Zhao, Armstrong, Darrell, Chung, Jain,
  Vanhoucke, Zhan, Zhou, Burgard, Chen, Wang, Zhu, Geng, Liu, Xu, Li, Lu, Ma,
  Kim, Chebotar, Zhou, Zhu, Wu, Xu, Wang, Bisk, Cho, Lee, Cui, Cao, Wu, Tang,
  Zhu, Zhang, Jiang, Li, Li, Iwasawa, Matsuo, Ma, Xu, Cui, Zhang, and
  Lin}]{DBLP:conf/icra/ONeillRMGPLPGMJ24}
Abby O'Neill, Abdul Rehman, Abhiram Maddukuri, Abhishek Gupta, Abhishek
  Padalkar, Abraham Lee, Acorn Pooley, Agrim Gupta, Ajay Mandlekar, Ajinkya
  Jain, Albert Tung, Alex Bewley, Alexander Herzog, Alex Irpan, Alexander
  Khazatsky, Anant Rai, Anchit Gupta, Andrew~E. Wang, Anikait Singh, and 260
  others. 2024.
\newblock Open x-embodiment: Robotic learning datasets and {RT-X} models : Open
  x-embodiment collaboration.
\newblock In \emph{{ICRA}}, pages 6892--6903. {IEEE}.

\bibitem[{Pan et~al.(2022)Pan, Zhu, Wang, and Yang}]{DBLP:conf/nips/Pan0WY22}
Minting Pan, Xiangming Zhu, Yunbo Wang, and Xiaokang Yang. 2022.
\newblock Iso-dream: Isolating and leveraging noncontrollable visual dynamics
  in world models.
\newblock In \emph{NeurIPS}.

\bibitem[{Radford et~al.(2021)Radford, Kim, Hallacy, Ramesh, Goh, Agarwal,
  Sastry, Askell, Mishkin, Clark, Krueger, and
  Sutskever}]{DBLP:conf/icml/RadfordKHRGASAM21}
Alec Radford, Jong~Wook Kim, Chris Hallacy, Aditya Ramesh, Gabriel Goh,
  Sandhini Agarwal, Girish Sastry, Amanda Askell, Pamela Mishkin, Jack Clark,
  Gretchen Krueger, and Ilya Sutskever. 2021.
\newblock Learning transferable visual models from natural language
  supervision.
\newblock In \emph{{ICML}}, volume 139 of \emph{Proceedings of Machine Learning
  Research}, pages 8748--8763. {PMLR}.

\bibitem[{Radosavovic et~al.(2022)Radosavovic, Xiao, James, Abbeel, Malik, and
  Darrell}]{DBLP:conf/corl/RadosavovicXJAM22}
Ilija Radosavovic, Tete Xiao, Stephen James, Pieter Abbeel, Jitendra Malik, and
  Trevor Darrell. 2022.
\newblock Real-world robot learning with masked visual pre-training.
\newblock In \emph{CoRL}, volume 205 of \emph{Proceedings of Machine Learning
  Research}, pages 416--426. {PMLR}.

\bibitem[{Raffel et~al.(2020)Raffel, Shazeer, Roberts, Lee, Narang, Matena,
  Zhou, Li, and Liu}]{DBLP:journals/jmlr/RaffelSRLNMZLL20}
Colin Raffel, Noam Shazeer, Adam Roberts, Katherine Lee, Sharan Narang, Michael
  Matena, Yanqi Zhou, Wei Li, and Peter~J. Liu. 2020.
\newblock Exploring the limits of transfer learning with a unified text-to-text
  transformer.
\newblock \emph{J. Mach. Learn. Res.}, 21:140:1--140:67.

\bibitem[{Reed et~al.(2022)Reed, Zolna, Parisotto, Colmenarejo, Novikov,
  Barth{-}Maron, Gimenez, Sulsky, Kay, Springenberg, Eccles, Bruce, Razavi,
  Edwards, Heess, Chen, Hadsell, Vinyals, Bordbar, and
  de~Freitas}]{DBLP:journals/tmlr/ReedZPCNBGSKSEBREHCHVBF22}
Scott~E. Reed, Konrad Zolna, Emilio Parisotto, Sergio~G{\'{o}}mez Colmenarejo,
  Alexander Novikov, Gabriel Barth{-}Maron, Mai Gimenez, Yury Sulsky, Jackie
  Kay, Jost~Tobias Springenberg, Tom Eccles, Jake Bruce, Ali Razavi, Ashley
  Edwards, Nicolas Heess, Yutian Chen, Raia Hadsell, Oriol Vinyals, Mahyar
  Bordbar, and Nando de~Freitas. 2022.
\newblock A generalist agent.
\newblock \emph{Trans. Mach. Learn. Res.}, 2022.

\bibitem[{Robine et~al.(2023)Robine, H{\"{o}}ftmann, Uelwer, and
  Harmeling}]{DBLP:conf/iclr/RobineHUH23}
Jan Robine, Marc H{\"{o}}ftmann, Tobias Uelwer, and Stefan Harmeling. 2023.
\newblock Transformer-based world models are happy with 100k interactions.
\newblock In \emph{{ICLR}}. OpenReview.net.

\bibitem[{Song et~al.(2023{\natexlab{a}})Song, Zhang, and
  King}]{DBLP:conf/nips/SongZK23}
Zixing Song, Yifei Zhang, and Irwin King. 2023{\natexlab{a}}.
\newblock No change, no gain: Empowering graph neural networks with expected
  model change maximization for active learning.
\newblock In \emph{NeurIPS}.

\bibitem[{Song et~al.(2023{\natexlab{b}})Song, Zhang, and
  King}]{DBLP:conf/nips/SongZK23a}
Zixing Song, Yifei Zhang, and Irwin King. 2023{\natexlab{b}}.
\newblock Optimal block-wise asymmetric graph construction for graph-based
  semi-supervised learning.
\newblock In \emph{NeurIPS}.

\bibitem[{Stone et~al.(2023)Stone, Xiao, Lu, Gopalakrishnan, Lee, Vuong,
  Wohlhart, Zitkovich, Xia, Finn, and
  Hausman}]{DBLP:journals/corr/abs-2303-00905}
Austin Stone, Ted Xiao, Yao Lu, Keerthana Gopalakrishnan, Kuang{-}Huei Lee,
  Quan Vuong, Paul Wohlhart, Brianna Zitkovich, Fei Xia, Chelsea Finn, and
  Karol Hausman. 2023.
\newblock Open-world object manipulation using pre-trained vision-language
  models.
\newblock \emph{CoRR}, abs/2303.00905.

\bibitem[{Sun et~al.(2023)Sun, Ma, Madaan, Bonatti, Huang, and
  Kapoor}]{DBLP:conf/iclr/SunMMBHK23}
Yanchao Sun, Shuang Ma, Ratnesh Madaan, Rogerio Bonatti, Furong Huang, and
  Ashish Kapoor. 2023.
\newblock {SMART:} self-supervised multi-task pretraining with control
  transformers.
\newblock In \emph{{ICLR}}. OpenReview.net.

\bibitem[{Tao et~al.(2024)Tao, Xiang, Shukla, Qin, Hinrichsen, Yuan, Bao, Lin,
  Liu, Chan, Gao, Li, Mu, Xiao, Gurha, Huang, Calandra, Chen, Luo, and
  Su}]{DBLP:journals/corr/abs-2410-00425}
Stone Tao, Fanbo Xiang, Arth Shukla, Yuzhe Qin, Xander Hinrichsen, Xiaodi Yuan,
  Chen Bao, Xinsong Lin, Yulin Liu, Tse{-}kai Chan, Yuan Gao, Xuanlin Li,
  Tongzhou Mu, Nan Xiao, Arnav Gurha, Zhiao Huang, Roberto Calandra, Rui Chen,
  Shan Luo, and Hao Su. 2024.
\newblock Maniskill3: {GPU} parallelized robotics simulation and rendering for
  generalizable embodied {AI}.
\newblock \emph{CoRR}, abs/2410.00425.

\bibitem[{van~den Oord et~al.(2018)van~den Oord, Li, and
  Vinyals}]{DBLP:journals/corr/abs-1807-03748}
A{\"{a}}ron van~den Oord, Yazhe Li, and Oriol Vinyals. 2018.
\newblock Representation learning with contrastive predictive coding.
\newblock \emph{CoRR}, abs/1807.03748.

\bibitem[{Vaswani et~al.(2017)Vaswani, Shazeer, Parmar, Uszkoreit, Jones,
  Gomez, Kaiser, and Polosukhin}]{DBLP:conf/nips/VaswaniSPUJGKP17}
Ashish Vaswani, Noam Shazeer, Niki Parmar, Jakob Uszkoreit, Llion Jones,
  Aidan~N. Gomez, Lukasz Kaiser, and Illia Polosukhin. 2017.
\newblock Attention is all you need.
\newblock In \emph{{NIPS}}, pages 5998--6008.

\bibitem[{Vemprala et~al.(2023)Vemprala, Bonatti, Bucker, and
  Kapoor}]{DBLP:journals/corr/abs-2306-17582}
Sai Vemprala, Rogerio Bonatti, Arthur Bucker, and Ashish Kapoor. 2023.
\newblock Chatgpt for robotics: Design principles and model abilities.
\newblock \emph{CoRR}, abs/2306.17582.

\bibitem[{Wu et~al.(2024)Wu, Jing, Cheang, Chen, Xu, Li, Liu, Li, and
  Kong}]{DBLP:conf/iclr/WuJCCXLLLK24}
Hongtao Wu, Ya~Jing, Chilam Cheang, Guangzeng Chen, Jiafeng Xu, Xinghang Li,
  Minghuan Liu, Hang Li, and Tao Kong. 2024.
\newblock Unleashing large-scale video generative pre-training for visual robot
  manipulation.
\newblock In \emph{{ICLR}}. OpenReview.net.

\bibitem[{Xiao et~al.(2025)Xiao, Liu, Wang, Zhou, Qi, Jiang, He, and
  Cheng}]{DBLP:journals/ijon/XiaoLWZQJHC25}
Xuan Xiao, Jiahang Liu, Zhipeng Wang, Yanmin Zhou, Yong Qi, Shuo Jiang, Bin He,
  and Qian Cheng. 2025.
\newblock \href {https://doi.org/10.1016/J.NEUCOM.2025.129963} {Robot learning
  in the era of foundation models: a survey}.
\newblock \emph{Neurocomputing}, 638:129963.

\bibitem[{Xu et~al.(2024)Xu, Wu, Wen, Li, Liu, Che, and
  Tang}]{DBLP:journals/corr/abs-2402-02385}
Zhiyuan Xu, Kun Wu, Junjie Wen, Jinming Li, Ning Liu, Zhengping Che, and Jian
  Tang. 2024.
\newblock \href {https://doi.org/10.48550/ARXIV.2402.02385} {A survey on
  robotics with foundation models: toward embodied {AI}}.
\newblock \emph{CoRR}, abs/2402.02385.

\bibitem[{Zhang et~al.(2024{\natexlab{a}})Zhang, Huang, Jin, and
  Lu}]{DBLP:journals/pami/ZhangHJL24}
Jingyi Zhang, Jiaxing Huang, Sheng Jin, and Shijian Lu. 2024{\natexlab{a}}.
\newblock Vision-language models for vision tasks: {A} survey.
\newblock \emph{{IEEE} Trans. Pattern Anal. Mach. Intell.}, 46(8):5625--5644.

\bibitem[{Zhang et~al.(2024{\natexlab{b}})Zhang, Zhu, Song, Chen, Fu, Meng,
  Koniusz, and King}]{DBLP:conf/kdd/ZhangZS00MKK24}
Yifei Zhang, Hao Zhu, Zixing Song, Yankai Chen, Xinyu Fu, Ziqiao Meng, Piotr
  Koniusz, and Irwin King. 2024{\natexlab{b}}.
\newblock Geometric view of soft decorrelation in self-supervised learning.
\newblock In \emph{{KDD}}, pages 4338--4349. {ACM}.

\bibitem[{Zhao et~al.(2023)Zhao, Kumar, Levine, and
  Finn}]{DBLP:conf/rss/ZhaoKLF23}
Tony~Z. Zhao, Vikash Kumar, Sergey Levine, and Chelsea Finn. 2023.
\newblock Learning fine-grained bimanual manipulation with low-cost hardware.
\newblock In \emph{Robotics: Science and Systems}.

\end{thebibliography}
